\documentclass[lettersize,journal]{IEEEtran}
\usepackage{amsmath,amsfonts}
\usepackage{algorithmic}
\usepackage{algorithm}
\usepackage{array}
\usepackage{textcomp}
\usepackage{stfloats}
\usepackage{url}
\usepackage{verbatim}
\usepackage{graphicx}
\usepackage{cite}
\usepackage{subfigure}
\usepackage{epsfig}
\usepackage{amssymb}
\usepackage{booktabs}
\usepackage{colortbl}
\usepackage{xcolor}
\usepackage{multirow}
\usepackage{pifont}
\usepackage{soul}
\usepackage{paralist}
\usepackage[citecolor=blue, colorlinks]{hyperref}
\usepackage[normalem]{ulem}
\usepackage{enumitem}
\usepackage{threeparttable}
\definecolor{light-gray}{gray}{0.95}

\begin{document}

\title{PointVST: Self-Supervised Pre-training for \\ 3D Point Clouds via View-Specific \\ Point-to-Image Translation}

\author{Qijian Zhang and Junhui Hou, \textit{Senior Member}, \textit{IEEE}
\thanks{This project was supported by the Hong Kong Research Grants Council
under Grant 11219422 and Grant 11202320.}
\thanks{The authors are with the Department of Computer Science, City University of Hong Kong, Hong Kong SAR.}
}

\markboth{}
{Shell \MakeLowercase{\textit{et al.}}: A Sample Article Using IEEEtran.cls for IEEE Journals}

\maketitle

\begin{abstract}
    The past few years have witnessed the great success and prevalence of self-supervised representation learning within the language and 2D vision communities. However, such advancements have not been fully migrated to the field of 3D point cloud learning. Different from existing pre-training paradigms designed for deep point cloud feature extractors that fall into the scope of generative modeling or contrastive learning, this paper proposes a translative pre-training framework, namely PointVST, driven by a novel self-supervised pretext task of cross-modal translation from 3D point clouds to their corresponding diverse forms of 2D rendered images. More specifically, we begin with deducing view-conditioned point-wise embeddings through the insertion of the viewpoint indicator, and then adaptively aggregate a view-specific global codeword, which can be further fed into subsequent 2D convolutional translation heads for image generation. Extensive experimental evaluations on various downstream task scenarios demonstrate that our PointVST shows consistent and prominent performance superiority over current state-of-the-art approaches as well as satisfactory domain transfer capability. Our code will be publicly available at \url{https://github.com/keeganhk/PointVST}.
\end{abstract}


\section{Introduction} \label{sec:introduction}

\IEEEPARstart{A}{s} one of the most straightforward 3D data representation modalities faithfully depicting the raw information of the target geometric structures, point clouds have been playing a critical role in a wide range of real-world application scenarios such as immersive telepresence, autonomous driving, SLAM, and robotics.

In recent years, empowered by the continuous progress of deep set architecture design \cite{qi2017pointnet,qi2017pointnet++,li2018pointcnn,wang2019dynamic,liu2019relation,thomas2019kpconv,xiang2021walk,zhao2021point}, deep learning-based frameworks that directly work on 3D point clouds have been richly investigated and applied for various downstream task scenarios of low-level geometric processing \cite{yu2018pu,wang2019deep,aoki2019pointnetlk} and high-level semantic understanding \cite{uy2018pointnetvlad,shi2019pointrcnn,hu2020randla,zhang2023pointmcd}. Essentially, these approaches rely on large-scale annotated point cloud repositories \cite{geiger2012we,armeni20163d,mo2019partnet} to achieve competitive performances in a supervised learning manner. However, manually annotating massive amounts of irregular and unstructured 3D data is known to be laborious and cumbersome, which can restrict the broad applicability of 3D point cloud representations.

Fortunately, the thriving development and rapid popularization of 3D scanning technologies (e.g., LiDAR, Kinect, stereo cameras) facilitate convenient construction of large-scale \textit{unlabeled} 3D point cloud repositories, which strongly motivate a recent line of works \cite{sauder2019self,poursaeed2020self,sharma2020self,wang2021unsupervised,xie2020pointcontrast,huang2021spatio,zhang2021self,afham2022crosspoint} exploring self-supervised pre-training approaches for point cloud representation learning without manual annotations. Functionally, an effective pre-training process enhances the robustness and generalization ability of backbone encoders by guiding the learning of generic and transferable features, and thus reduces the amount of annotated data required for task-specific supervised learning. Still, despite the remarkable success of self-supervised pre-training within the language and 2D vision communities \cite{devlin2018bert,jing2020self}, its potential has not been fully realized and explored in the fast-growing area of 3D point cloud learning.

\begin{figure}[t]
	\centering
	\includegraphics[width=1.0\linewidth]{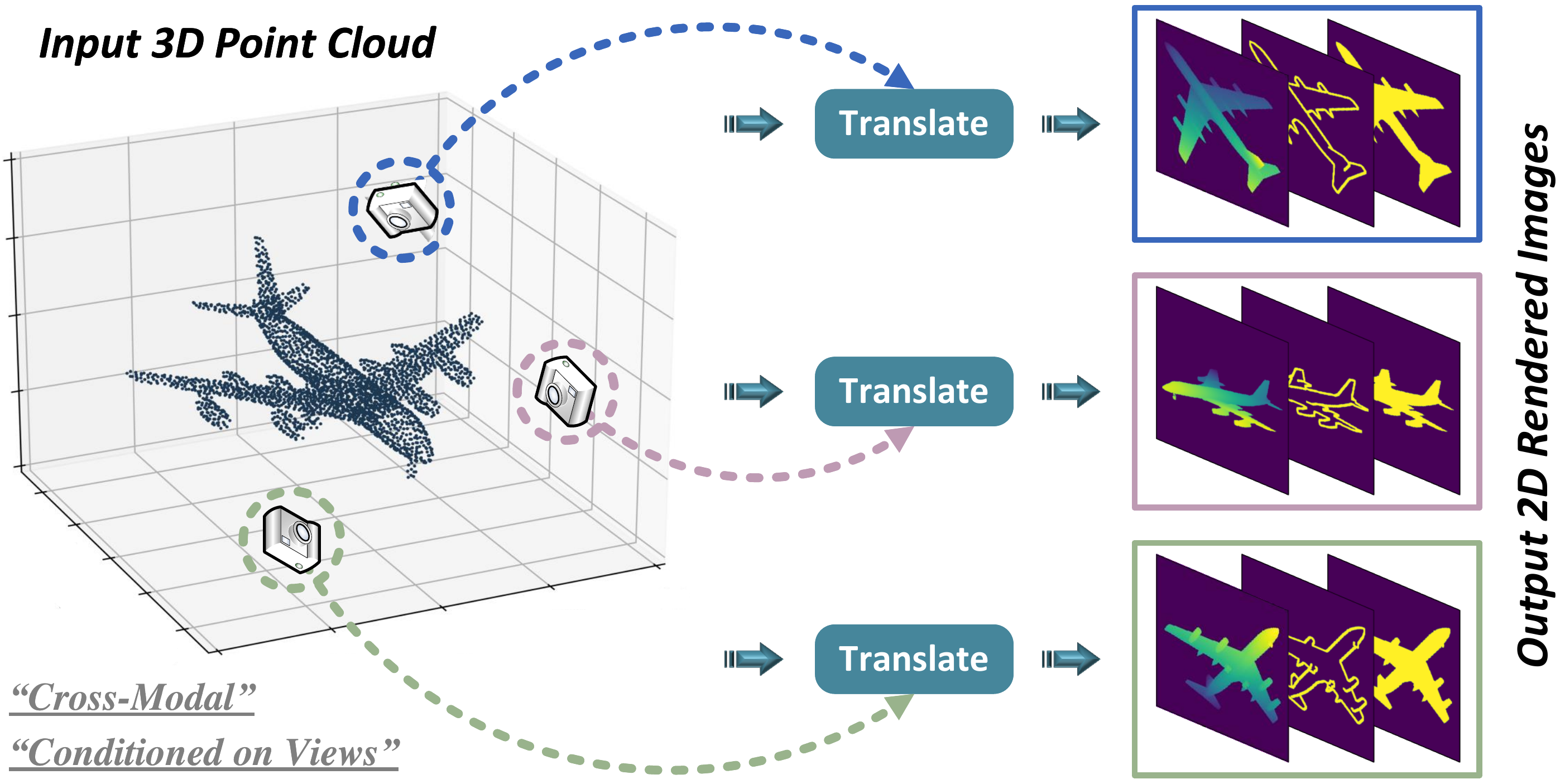}
	\caption{\textbf{Illustration of view-specific cross-modal translation.} For an input 3D point cloud and an arbitrarily specified camera position, we extract view-specific point cloud feature embeddings, from which we further generate the corresponding diverse forms of 2D rendered images (e.g., depth, contour, and silhouette maps). Such a self-supervised pretext task can effectively drive the learning of discriminative and transferable point cloud feature representations during the pre-training phase.}
	\vspace{-0.3cm}
	\label{fig:idea-illustration}
\end{figure}

Basically, the core of self-supervised learning lies in the design of appropriate pretext tasks and the specific implementation of its learning process. Therefore, depending on the difference of pretext tasks \cite{liu2021self}, existing self-supervised pre-training frameworks customized for point clouds can be broadly categorized into \textit{generative modeling} \cite{sauder2019self,poursaeed2020self,sharma2020self,wang2021unsupervised} and \textit{contrastive learning} \cite{sanghi2020info3d,xie2020pointcontrast,zhang2021self,du2021self,huang2021spatio,afham2022crosspoint}. Architecturally, the former generative paradigms are typically implemented as self-reconstructive pipelines, i.e., the input point cloud is encoded in the latent space to produce feature embeddings, from which the original 3D geometry information is inferred/restored. Still, these methods show the relatively limited capability of driving the backbone encoder to learn discriminative representations. As revealed in \cite{sauder2019self,zeng2021corrnet3d,feng2021recurrent}, the reasons could be attributed to the possibly-problematic point set similarity metrics such as Chamfer distance (CD) and earth mover's distance (EMD) that are hard to optimize during generation, and the less powerful point-tailored modeling components built upon shared multi-layer perceptrons (MLPs). The latter contrastive paradigms aim to learn feature space similarity measurements between positive and negative samples created by imposing different transformations on the given point cloud instances. These approaches show better performances and gain increasingly growing attention. Still, one may have to carefully design the network components and cautiously implement the actual parameter optimization and updating process, in order to circumvent the typical issue of model collapse \cite{grill2020bootstrap,jing2022understanding}.

Different from existing generative or contrastive paradigms, this paper poses a new perspective in designing self-supervised pre-training frameworks for 3D point clouds via innovatively introducing a novel translative pretext task, namely PointVST. As shown in Figure~\ref{fig:idea-illustration}, the basic idea is to perform cross-modal translation conditioned on views from a 3D point cloud to its diverse forms of 2D rendered images (e.g., depth, contour, and silhouette maps), instead of the given point cloud itself. \textit{The conversion of generation objectives from the original irregular 3D geometry space to the regular 2D image domain} can make a big difference to the overall learning effects while naturally circumventing the typical limitations of generative paradigms, as elaborated in the following aspects:
\begin{itemize}[leftmargin=10.5pt, topsep=1.0pt]
	\item[--] We are able to get rid of the point set similarity metrics (CD and EMD), which are proven to be weak owing to unknown correspondences between ground-truth and generated point clouds \cite{sauder2019self,zeng2021corrnet3d,feng2021recurrent} and hence result in limited feature representation capability. Naturally, we can directly impose stronger supervisions by minimizing pixel-wise errors between ground-truth images and the translated results, which can be much easier to optimize.
	\item[--] We can resort to mature and powerful 2D CNNs, rather than relatively less expressive point-tailored network modules, to implement the reconstruction process. Overall, as a different manner of achieving recovery of 3D geometric information, our proposed cross-modal translative paradigm can subtly circumvent several aspects of challenging problems induced by straightforward point cloud reconstruction.
	\item[--] Besides, the view-specific translation mechanism facilitates more fine-grained feature representation learning of geometric structures.
\end{itemize}

In practice, we conduct extensive experiments over a variety of point cloud processing tasks, including shape classification, object part segmentation, normal estimation, and scene semantic segmentation. With diverse evaluation protocols, PointVST shows prominent performance superiorities over current state-of-the-art approaches, indicating the great potential of the new translative pre-training paradigm.

The remainder of this paper is organized as follows. Section II systematically reviews influential point cloud representation learning approaches and the recent self-supervised pre-training frameworks. In Section III, we introduce our proposed PointVST workflow for pre-training deep point cloud feature extractors. Section IV provides comprehensive experimental verifications, performance comparisons, and ablation studies to demonstrate the effectiveness and superiority of our approach. In the end, we conclude the whole paper in Section V.

\section{Related Works} \label{sec:related-works}

\subsection{Point Cloud Representation Learning} \label{sec:pc-rl}

Unlike conventional 2D representation learning dealing with structured visual signals (images, videos) defined on dense and regular grids, 3D point clouds are characterized by irregularity and unorderedness. Such unstructured nature poses significant difficulties in learning discriminative geometric feature representations. Pioneered by Qi \textit{et al}. \cite{qi2017pointnet,qi2017pointnet++}, there have emerged numerous deep set architectures \cite{li2018pointcnn,wang2019dynamic,thomas2019kpconv,wu2019pointconv,liu2019relation,xu2020grid,xu2021paconv,xiang2021walk} that directly operate on raw point sets without any pre-processing. 

To overcome the dependence on large-scale annotated point cloud data, there have been a plethora of works focusing on performing unsupervised representation learning on unlabeled point clouds built upon various paradigms of self-reconstructive learning pipelines, such as auto-encoders (AEs) \cite{li2018so,yang2018foldingnet,chen2019deep,hassani2019unsupervised,zhao20193d,liu2019l2g,han2019multi}, generative adversarial networks (GANs) \cite{wu2016learning,valsesia2018learning,achlioptas2018learning,han2019view}, and others \cite{yang2019pointflow,sun2020pointgrow}. However, the actual feature learning efficacy of these unsupervised approaches is still far from satisfactory.

\begin{figure*}[t]
	\centering
	\includegraphics[width=1.0\linewidth]{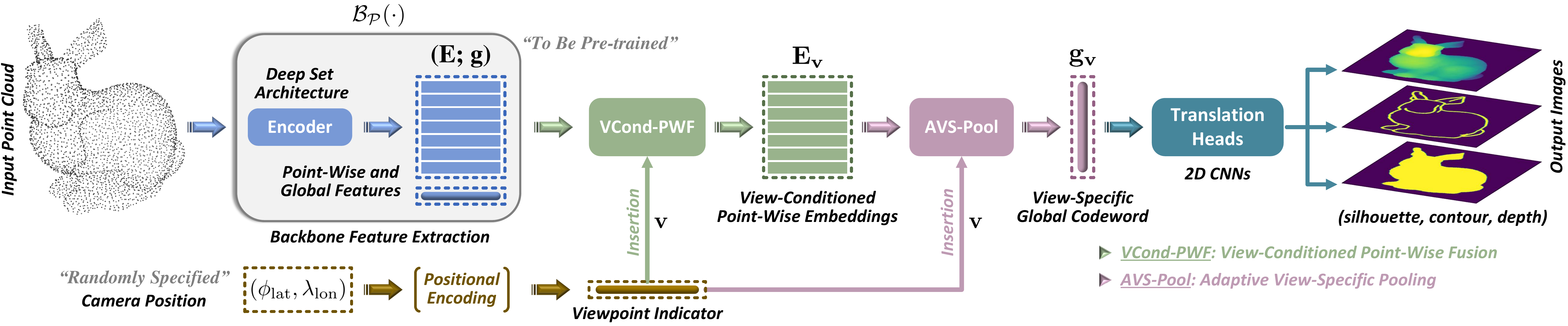}
	\caption{\textbf{Overall workflow of the proposed PointVST for self-supervised point cloud pre-training.} Given an input 3D point cloud, we can deduce from the target backbone encoder $\mathcal{B_P}(\cdot)$ a set of point-wise features $\mathbf{E}$ and a global codeword $\mathbf{g}$. Within the pretext task, given a randomly specified camera position, we produce a viewpoint indicator $\mathbf{v}$ and fuse it with backbone features to obtain view-conditioned point-wise embeddings $\mathbf{E_v}$. After that, we customize AVS-Pool to aggregate from $\mathbf{E_v}$ a view-specific global codeword $\mathbf{g_v}$, which encodes partial 3D geometric patterns exactly corresponding to the viewpoint condition. Finally, we feed $\mathbf{g_v}$ into the subsequent translation heads built upon standard 2D CNNs to generate different forms of rendered images.}
	\label{fig:overall-workflow}
\end{figure*}

\subsection{Self-Supervised Point Cloud Pre-training} \label{sec:pc-ssp}

More recently, self-supervised learning has been adapted for point cloud pre-training pipelines, demonstrating dominating performances on various shape analysis and scene understanding tasks. Generally, previous works can be categorized into generative modeling and contrastive learning paradigms.

\textbf{\textit{Generative Paradigms}.} Based on the intuition that effective feature abstractions should contain sufficient information to reconstruct the original geometric structures, a family of generative pre-training pipelines has been proposed by designing various appropriate pretext tasks. Jigsaw~\cite{sauder2019self} recovers point clouds whose parts have been randomly rearranged. OrienEst~\cite{poursaeed2020self} tends to predict 3D rotation angles, motivated by the preceding practice for image learning \cite{komodakis2018unsupervised}. GLBR~\cite{rao2020global} performs bidirectional reasoning between the global feature and different scales of local representations. cTree~\cite{sharma2020self} hierarchically organizes input point cloud data in a cover-tree and predicts the decomposition patterns. OcCo~\cite{wang2021unsupervised} constructs a simple but highly effective approach based on partial point cloud completion.

\textbf{\textit{Contrastive Paradigms}.} The core of contrastive learning lies in the feature similarity measurement between positive and negative samples. Info3D~\cite{sanghi2020info3d} takes inspiration from \cite{oord2018representation,velickovic2019deep} to maximize the mutual information between a complete 3D object and local chunks as well as its geometrically transformed version. PointContrast~\cite{xie2020pointcontrast} aims to learn invariant point-level feature mappings between transformed views of given point cloud scenes. DepthContrast~\cite{zhang2021self} deals with single-view depth scans without the requirement of 3D registration and point-wise correspondences by performing instance discrimination after global feature pooling. SelfContrast~\cite{du2021self} proposes to contrast patches within complete point clouds for mining non-local self-similarity relationships. As an extension of BYOL \cite{grill2020bootstrap}, STRL~\cite{huang2021spatio} learns discriminative point cloud features via interactions of the online and target networks. CrossPoint~\cite{afham2022crosspoint} adopts a multi-modal learning framework composed of a 3D point cloud branch and a textured 2D image branch, whose pre-training objective is jointly driven by the intra-modal invariance between different point cloud augmentations and cross-modal contrastive constraints between rendered image features and point cloud prototype features. MvDeCor~\cite{sharma2022mvdecor} renders a given 3D shape from multiple views and uses a CNN backbone to produce pixel-wise embeddings, and the pre-training is driven by multi-view dense correspondence learning.

More recently, inspired by the success of masked auto-encoders (MAE) \cite{he2022masked} in the 2D vision community, some researchers are devoted to adapting MAE-like pre-training pipelines \cite{yu2022point,pang2022masked,liu2022masked,zhang2022pointmae}, which achieve impressive performances. These approaches are particularly customized for pre-training transformer-style point cloud backbones \cite{guo2021pct,zhao2021point}, and thus cannot be applied to other common types of deep set architectures. By contrast, our method and the above-discussed works pursue wider applicability to common types of point cloud backbone architectures.

\section{Proposed Method} \label{sec:proposed-method}

\subsection{Overview}

Given a specific backbone point cloud feature encoder $\mathcal{B_P}(\cdot)$, we aim at designing an appropriate self-supervised pretext task $\mathcal{T}(\cdot)$, which can effectively facilitate learning expressive and transferable 3D geometric representations from massive unlabeled point cloud data.

Formally, denote by $\mathbf{P} \in \mathbb{R}^{N \times 3}$ an input 3D point cloud that consists of $N$ spatial points $\{ \mathbf{x}_n \in \mathbb{R}^3 \}_{n=1}^{N}$. Without loss of generality, we can correspondingly obtain at the output end of $\mathcal{B_P}(\cdot)$ a set of point-wise features $\mathbf{E} \in \mathbb{R}^{N \times D_e}$ embedded in the high-dimensional latent space, as well as a vectorized codeword $\mathbf{g} \in \mathbb{R}^{D_g}$ serving as the global shape signature. Technically, as illustrated in Figure~\ref{fig:overall-workflow}, the overall workflow of our PointVST pre-training framework comprises three major stages stacked in an end-to-end manner. In the very beginning, given a randomly specified camera position, we produce via positional encoding a vectorized viewpoint indicator denoted as $\mathbf{v} \in \mathbb{R}^{D_i}$, which is further fused with raw backbone features ($\mathbf{E}$ and $\mathbf{g}$) to generate a set of view-conditioned point-wise embeddings denoted as $\mathbf{E_v} \in \mathbb{R}^{N \times D_v}$. After that, we customize an adaptive view-specific pooling (AVS-Pool) mechanism to aggregate from $\mathbf{E_v}$ a view-specific global codeword represented by $\mathbf{g_v}$. Intuitively, different from $\mathbf{g}$ that encodes the whole geometric structure of $\mathbf{P}$ conveyed by \textit{complete} points, $\mathbf{g_v}$ is supposed to depict \textit{partial} 3D structural information in the projection space that exactly corresponds to its viewpoint position. In the end, $\mathbf{g_v}$ passes through the subsequent translation heads that are built upon 2D CNNs to generate different forms of rendered images.

After finishing pre-training, the backbone encoder $\mathcal{B_P}(\cdot)$ can be directly applied to obtain generic and discriminative point cloud representations, or integrated into various task-specific learning pipelines for downstream fine-tuning.

\subsection{View-Conditioned Point-Wise Fusion}

By default, the input point cloud model has already been centralized and normalized into a unit sphere, such that we can configure a fixed observation distance (from the camera position to the object centroid). Thus, under the geographic coordinate system, we can describe any camera position by two parameters, i.e., latitude angle $\phi_{\mathrm{lat}} \in [-90^{\circ}, 90^{\circ}]$, and longitude angle $\lambda_{\mathrm{lon}} \in [0, 360^{\circ})$. Following previous practice \cite{mildenhall2020nerf} involving view-conditioned reasoning, instead of directly using raw latitude and longitude angles $(\phi_{\mathrm{lat}},\lambda_{\mathrm{lon}})$ as the condition signal, we perform positional encoding via several learnable transformation layers to enhance flexibility. More concretely, we separately lift $\phi_{\mathrm{lat}}$ and $\lambda_{\mathrm{lon}}$ into the latent space and then concatenate the resulting two high-dimensional vectors to deduce the viewpoint indicator:
\begin{equation}
	\mathbf{v} = \mathcal{F}_{\phi}(\phi_{\mathrm{lat}}) \oplus \mathcal{F}_{\lambda}(\lambda_{\mathrm{lon}}),
\end{equation}
where $\mathcal{F}_{\phi}$ and $\mathcal{F}_{\lambda}$ denote non-linear transformations implemented as two consecutive fully-connected (FC) layers, and $\oplus$ represents feature channel concatenation.

After that, we fuse raw backbone features and the viewpoint indicator to produce a set of view-specific point-wise embeddings $\mathbf{E_v}$, which can be formulated as:
\begin{equation}
	\mathbf{E_v} = \mathcal{M}_f(\mathcal{M}_e(\mathbf{E}) \oplus \mathcal{F}_g(\mathbf{g}) \oplus \mathbf{v}),
\end{equation}
\noindent where $\mathcal{M}_e$ and $\mathcal{M}_f$ are both implemented as shared MLPs, and $\mathcal{F}_g$ denotes a single FC layer. Note that all the embedding values in $\mathbf{E_v}$ tend to be non-negative, since we choose ReLU as the activation function.

\begin{figure}[t]
	\centering
	\includegraphics[width=1.0\linewidth]{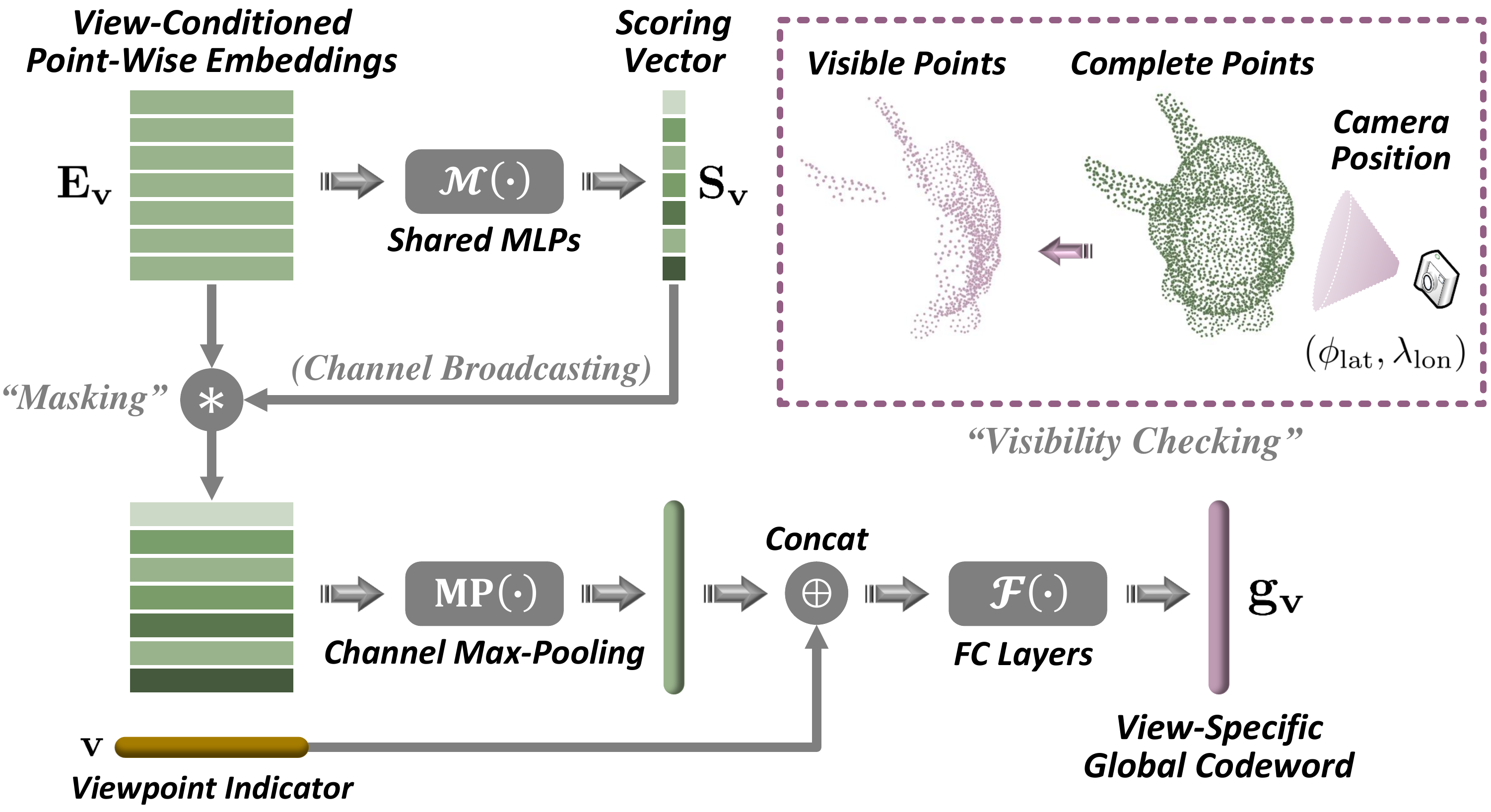}
	\caption{\textbf{Illustration of AVS-Pool, which is customized for aggregating view-conditioned point-wise embeddings $\mathbf{E_v}$ into a vectorized view-specific global codeword $\mathbf{g_v}$.}}
	\label{fig:avs-pool-workflow}
\end{figure}

\subsection{Adaptive View-Specific Pooling}

In the preceding stage, we insert the viewpoint cues in a fine-grained manner to obtain view-conditioned point-wise embeddings $\mathbf{E_v}$. In fact, when observing the target shape at the specified viewpoint, only a subset of \textit{visible} points in $\mathbf{P}$ can convey effective information in the projection space of the 2D viewing plane. Therefore, to facilitate view-specific learning, we expect that features of invisible points could be adaptively omitted from the subsequent pooling operation applied on $\mathbf{E_v}$. To achieve this goal, we investigate adaptive view-specific pooling (AVS-Pool) that aggregates $\mathbf{E_v}$ into a  view-specific global codeword $\mathbf{g_v}$, as shown in Figure~\ref{fig:avs-pool-workflow}.

Specifically, we start by explicitly predicting a positive scoring vector $\mathbf{S_v} \in \mathbb{R}^{N}$, serving as a point-wise visibility mask, by feeding $\mathbf{E_v}$ into shared MLPs with Sigmoid as the activation function at the output end, i.e.,
\begin{equation}
	\mathbf{S_v} = \sigma(\mathcal{M}_s(\mathbf{E_v})),
\end{equation}
where $\mathcal{M}_s$ denotes three consecutive layers of shared MLPs whose final layer is implemented as a linear transformation, and $\sigma(\cdot)$ denotes the Sigmoid function. Thus, each value of $\mathbf{S_v}$ is within the range of $(0, 1)$ and designed to indicate the visibility status of the corresponding point.

To guide the prediction of $\mathbf{S_v}$ during the whole pre-training process, we accordingly introduce a visibility constraint:
\begin{equation}
	\mathcal{C}_v = \lVert \mathbf{S_v} - \mathbf{S^{\prime}_v} \rVert_1,
\end{equation}
where $\mathbf{S^{\prime}_v} \in \{ 0, 1 \}$ denotes the ground-truth binary visibility mask ($1$ for visible, $0$ for invisible) acquired by applying point set visibility checking \cite{katz2007direct} on input points $\mathbf{P}$. Intuitively, the learning process of visibility prediction serves as an additional driving task to promote the view awareness of $\mathbf{E_v}$, which can bring better pre-training effects than directly using the ground-truth binary visibility mask.

By multiplying the predicted $\mathbf{S_v}$ with $\mathbf{E_v}$, we achieve the effects of ``masking'' for suppressing the feature responses of invisible points, and then apply channel max-pooling on the masked point-wise embeddings. After that, we concatenate the pooled vector with the viewpoint indicator, and further deploy FC layers to generate the view-specific global codeword $\mathbf{g_v}$, which can be formulated as
\begin{equation}
	\mathbf{g_v} = \mathcal{F}_s(\texttt{MP}(\mathbf{S_v} \ast \mathbf{E_v}) \oplus \mathbf{v}),
\end{equation}
where $\ast$ denotes element-wise multiplication with default channel-wise broadcasting, $\texttt{MP}(\cdot)$ represents channel-wise max-pooling, and $\mathcal{F}_s$ denotes two consecutive FC layers.

\subsection{Translation Heads and Loss Functions}

Inheriting conventional image auto-encoders, we start by reshaping $\mathbf{g_v}$ from the original vectorized representation to a $C_r$-dimensional 2D feature map of dimensions $H_r \times W_r$. After that, we deploy 2D CNNs as translation heads, composed of several stages of deconvolutional layers accompanied by progressive spatial up-scaling, to produce a higher-resolution feature map. Finally, we feed this 2D feature map into three parallel convolutional blocks to reconstruct three forms of 2D rendered images, which are chosen as 1) depth map $\mathbf{I}_d$; 2) silhouette mask $\mathbf{I}_s$; 3) contour map $\mathbf{I}_c$.

In principle, it is required that the rendering process of the image forms to be translated should purely depict the original 3D geometric properties and not interfered by extrinsic factors, such as texture, material, and illumination, meaning that RGB images may not be an optimal choice. Besides, we also expect that the preparation of ground-truth images as the supervision signals should be convenient, reliable, and efficient, in order to deal with sparse and noisy raw point clouds with huge data quantity. From this perspective, other more demanding image forms such as normal maps would become undesirable.

Practically, we start by rendering the ground-truth 2D depth map $\mathbf{I}^{\prime}_d$ from the specified camera pose. After that, we can deduce the ground-truth silhouette mask $\mathbf{I}^{\prime}_s$ and the ground-truth contour map $\mathbf{I}^{\prime}_c$ by applying binary thresholding and Canny edge detection on $\mathbf{I}^{\prime}_d$, respectively.

For supervision, we compute binary cross-entropy losses ($\mathcal{L}_s$ and $\mathcal{L}_c$) for the silhouette mask $\mathbf{I}_s$ and the contour map $\mathbf{I}_c$. For the regression of depth values, we compute the $L_1$ loss ($\mathcal{L}_d$) for the depth map $\mathbf{I}_d$. Thus, we can formulate the overall pre-training objective as
\begin{equation} \label{eqn-loss-overall}
	\mathcal{L}_\mathrm{overall} = \omega_v \mathcal{C}_v + \omega_d \mathcal{L}_d + \omega_s \mathcal{L}_s + \omega_c \mathcal{L}_c,
\end{equation}
where all weights $\omega_v$, $\omega_d$, $\omega_s$, and $\omega_c$ are set to 1 empirically.

\section{Experiments} \label{sec:exp}

We comprehensively validate the effectiveness and superiority of our proposed PointVST. In the following, we start by introducing necessary technical implementations, experimental setups, and evaluation principles in Sec.~\ref{sec:exp-implementation} and \ref{sec:exp-overall}. We present and compare downstream task performances from Sec.~\ref{sec:exp-obj-cls} to Sec.~\ref{sec:exp-sce-seg}. Subsequently, we conduct extensive ablation studies in Sec.~\ref{sec:ablation-study}. Finally, we re-emphasize our core contribution in Sec.~\ref{sec:disc}, while further pointing out potential limitations of the current technical implementations and future extensions.

\subsection{Implementation Details} \label{sec:exp-implementation}

\noindent \textbf{\textit{View-Conditioned Point-Wise Fusion}.} When specifying camera positions for 2D image rendering, we uniformly set the observation distance as $d=2$. We set the number of views per input point cloud as $8$, which is empirically found to achieve a reasonable trade-off between training costs and the stability and effects of the whole pre-training stage.

As presented in Eq.~(\textcolor{red}{1}) for obtaining the vectorized viewpoint indicator $\mathbf{v} \in \mathbb{R}^{D_i}$, both $\mathcal{F}_{\phi}$ and $\mathcal{F}_{\lambda}$ comprise two consecutive FC layers whose output channels are $\{64,128\}$. Thus, we uniformly have the dimension of the resulting viewpoint indicator $D_i=256$. After that, when performing point-wise fusion (Eq.~(\textcolor{red}{2})), the output channels of both $\mathcal{M}_e$ and $\mathcal{F}_g$ are $1024$, meaning that the input dimension of $\mathcal{M}_f$ is $1024+1024+256=2304$. And we set its output channels as $1024$, such that the dimension of the resulting view-conditioned point-wise embeddings $\mathbf{E_v}$ (i.e., $D_v$) is $1024$.
\vspace{0.25cm}

\noindent \textbf{\textit{Adaptive View-Specific Pooling}.} As formulated in Eq.~(\textcolor{red}{3}), AVS-Pool begins with the explicit prediction of the point-wise visibility scoring vector $\mathbf{S_v} \in \mathbb{R}^{N}$, where the output channels of the three-layer shared MLPs $\mathcal{M}_s$ are configured as $\{256,128,1\}$. To learn the visibility mask $\mathbf{S_v}$ in a supervised manner, we adopted a classic hidden point removal algorithm \cite{katz2007direct} developed in the graphics community, which directly operates on raw point sets while maintaining simple and fast. Applying this algorithm on the input point cloud $\mathbf{P}$ can conveniently generate the ground-truth mask $\mathbf{S^{\prime}_v}$ that indicates point-wise visibility status with respect to the specified viewpoint position with binary ($0$ or $1$) flags. After that, we can deduce the vectorized view-specific global codeword $\mathbf{g_v}$ following the procedures in  Eq.~(\textcolor{red}{5}), where the output channels of the two FC layers of $\mathcal{F}_s$ are $\{2048,2048\}$ (meaning that the dimension of $\mathbf{g_v}$ is $2048$).
\vspace{0.25cm}

\begin{table}[t]
	\centering	
	\renewcommand\arraystretch{1.05}
	\setlength{\tabcolsep}{20.0pt}
	\caption{\textbf{Linear SVM classification on ModelNet40.}}
	\begin{tabular}{ l c }
		\toprule[1.0pt]
		Method & OAcc \\
		\hline
		3D-GAN~\cite{wu2016learning} & 83.3 \\
		Latent-GAN~\cite{achlioptas2018learning} & 85.7 \\
		FoldingNet~\cite{yang2018foldingnet} & 88.4 \\
		PointCaps~\cite{zhao20193d} & 88.9 \\
		MTFL~\cite{hassani2019unsupervised} & 89.1 \\
		SelfContrast~\cite{du2021self} & 89.6 \\
		VIP-GAN~\cite{han2019view} & 90.2 \\
		\hline
		{[P]-Jigsaw~\cite{sauder2019self}} & 87.3 \\
		{[P]-OrienEst~\cite{poursaeed2020self}} & 88.6 \\
		{[P]-STRL~\cite{huang2021spatio}} & 88.3 \\
		{[P]-OcCo~\cite{wang2021unsupervised}} & 88.8 \\
		{[P]-CrossPoint~\cite{afham2022crosspoint}} & 89.1 \\
		\rowcolor{light-gray}
		{[P]-PointVST} & \textbf{90.2} \\
		\hline
		{[D]-Jigsaw~\cite{sauder2019self}} & 90.6 \\
		{[D]-OrienEst~\cite{poursaeed2020self}} & 90.8 \\
		{[D]-STRL~\cite{huang2021spatio}} & 90.9 \\
		{[D]-OcCo~\cite{wang2021unsupervised}} & 90.7 \\
		{[D]-CrossPoint~\cite{afham2022crosspoint}} & 91.2 \\
		\rowcolor{light-gray} 
		{[D]-PointVST} & \textbf{92.1} \\
		\bottomrule[1.0pt]
	\end{tabular}
	\label{tab:svm-cls--modelnet40}
\end{table}

\begin{table}[t]
	\centering	
	\renewcommand\arraystretch{1.05}
	\setlength{\tabcolsep}{16.0pt}
	\caption{\textbf{Linear SVM classification for handling inputs with z/z and SO3/SO3 rotations on ModelNet40.} For the z/z setting, each input is rotated around the ground-axis with a uniform interval of $15^{\circ}$. For the SO3/SO3 setting, each input is repeatedly rotated $16$ times by randomly sampling from the 3D rotation group SO(3).}
	\begin{tabular}{ l c c }
		\toprule[1.0pt]
		\multirow{2}{*}{Method} & \multicolumn{2}{c}{ModelNet40} \\
		\cline{2-3} 
		& z/z & SO3/SO3 \\
		\hline
		{[P]-OcCo~\cite{wang2021unsupervised}} & 83.9 & 50.6 \\
		{[P]-CrossPoint~\cite{afham2022crosspoint}} & 86.5 & 58.4 \\
		\rowcolor{light-gray}
		{[P]-PointVST} & \textbf{87.6} & \textbf{64.7} \\
		\hline
		{[D]-OcCo~\cite{wang2021unsupervised}} & 86.3 & 55.7 \\
		{[D]-CrossPoint~\cite{afham2022crosspoint}} & 88.2 & 74.8 \\
		\rowcolor{light-gray}
		{[D]-PointVST} & \textbf{90.0} & \textbf{78.3} \\		
		\bottomrule[1.0pt]
	\end{tabular}
	\label{tab:svm-cls-rot--modelnet40}
\end{table}

\noindent \textbf{\textit{2D Image Translation Heads}.} To feed $\mathbf{g_v}$ into 2D convolutional layers, we deployed a single FC layer to lift its number of channels to $8192$, then performed reshaping from the vectorized structure to a 2D feature map of dimensions $8 \times 32 \times 32$ ($C_r=8$, $H_r=W_r=32$). The subsequent translation head network is sequentially composed of a convolutional layer (with output channels of $32$), two residual convolutional blocks \cite{he2016deep} (with output channels of $64$ and $128$ respectively) each with $2 \times$ bilinear feature interpolation, and a convolutional layer (with output channels of $512$). Thus, the dimension of the resulting 2D feature map is $512 \times 128 \times 128$. Finally, to reconstruct the tree forms (i.e., depth, contour, and silhouette maps) of rendered images, we deployed three (non-shared) output blocks, each of which consists of three convolutional layers (with output channels of $\{128,64,1\}$ respectively and the Sigmoid activation function in the end).

After rendering the ground-truth depth map $\mathbf{I}^{\prime}_d$, we can straightforwardly deduce the binary ground-truth silhouette mask $\mathbf{I}^{\prime}_s$, i.e., all non-empty depth values are set as $1$, and can produce the ground-truth contour map $\mathbf{I}^{\prime}_c$ by applying Canny edge detection. Additionally, we also applied the classic morphological dilation operation on the binary edges to strengthen thin/broken structures.

\begin{table}[t]
	\centering	
	\renewcommand\arraystretch{1.05}
	\setlength{\tabcolsep}{8.0pt}
	\caption{\textbf{Real-scanned object classification on ScanObjectNN (OBJ-BG) by fine-tuning from pre-trained backbones.}}
	\begin{tabular}{ l c | l c }
		\toprule[1.0pt]
		Method & OAcc & Method & OAcc \\
		\hline
		{[P]-Random} & 73.3 & {[D]-Random} & 82.8 \\
		{[P]-OcCo~\cite{wang2021unsupervised}} & 79.8 & {[D]-OcCo~\cite{wang2021unsupervised}} & 84.5 \\
		{[P]-CrossPoint~\cite{afham2022crosspoint}} & 80.2 & {[D]-CrossPoint~\cite{afham2022crosspoint}} & 86.2 \\
		\rowcolor{light-gray}
		{[P]-PointVST} & \textbf{80.7} & {[D]-PointVST} & \textbf{89.3} \\
		\bottomrule[1.0pt]
	\end{tabular}
	\label{tab:ft-cls--scanobjectnn}
\end{table}

\begin{table}[t]
	\centering	
	\renewcommand\arraystretch{1.05}
	\setlength{\tabcolsep}{10.0pt}
	\caption{\textbf{Part segmentation on ShapeNetPart by fine-tuning from pre-trained backbones.}}
	\begin{tabular}{ c c | c c }
		\toprule[1.0pt]
		\multicolumn{1}{c}{Method} & mIoU & \multicolumn{1}{c}{Method} & mIoU \\
		\hline
		\multicolumn{1}{c}{{[}P{]}-Random} & 83.7 & \multicolumn{1}{c}{{[}D{]}-Random} & 85.1 \\
		\multicolumn{1}{c}{{[}P{]}-OcCo~\cite{wang2021unsupervised}} & 84.4 & \multicolumn{1}{c}{{[}D{]}-OcCo~\cite{wang2021unsupervised}} & 85.4 \\
		\multicolumn{1}{c}{{[}P{]}-CrossPoint~\cite{afham2022crosspoint}} & 85.0 & \multicolumn{1}{c}{{[}D{]}-CrossPoint~\cite{afham2022crosspoint}} & 85.5 \\
		\rowcolor{light-gray}
		\multicolumn{1}{c}{{[}P{]}-PointVST}   & \textbf{86.8}   & \multicolumn{1}{c}{{[}D{]}-PointVST}   & \textbf{87.4} \\
		\toprule[1.0pt]
		\multicolumn{2}{c|}{Method} & \multicolumn{2}{c}{mIoU} \\
		\hline
		\multicolumn{2}{c|}{{[SR-UNet]}-Random} & \multicolumn{2}{c}{84.7} \\
		\multicolumn{2}{c|}{{[SR-UNet]}-PointContrast~\cite{xie2020pointcontrast}} & \multicolumn{2}{c}{85.1} \\
		\bottomrule[1.0pt]
\end{tabular}
\label{tab:ft-seg--shapenetpart}
\end{table}

\subsection{Pre-training Setups and Evaluation Schemes} \label{sec:exp-overall}

As a large-scale 3D shape repository consisting of more than 50,000 object models covering 55 semantic categories, ShapeNet~\cite{chang2015shapenet} has been widely adopted as the source dataset for pre-training in many previous works (e.g., \cite{sauder2019self,poursaeed2020self,huang2021spatio,afham2022crosspoint,yu2022point,pang2022masked}), in which point clouds can be acquired by sampling from the original mesh models. We followed such common practice to perform pre-training on ShapeNet and evaluated on diverse downstream tasks, including object classification (Sec.~\ref{sec:exp-obj-cls}), part segmentation (Sec.~\ref{sec:exp-par-seg}), normal estimation (Sec.~\ref{sec:exp-nor-est}), and scene semantic segmentation (Sec.~\ref{sec:exp-sce-seg}). For works (e.g., \cite{wang2021unsupervised}) that originally adopted inconsistent pre-training data (e.g., the smaller ModelNet~\cite{wu20153d} dataset), we used their official code to finish the pre-training process on the same data prepared from ShapeNet. For data preparation on ShapeNet~\cite{chang2015shapenet}, we used the commonly-adopted Poisson disk sampling algorithm to generate point cloud data (without auxiliary attributes) from the original polygon mesh models. The number of points is set as $2048$ for this setting. During the rendering process, the resulting image resolution is configured as $128 \times 128$, which can effectively maintain necessary geometric details while achieving satisfactory pre-training efficiency.

\begin{table}[t]
	\centering
	\renewcommand\arraystretch{1.05}
	\setlength{\tabcolsep}{8.0pt}
	\caption{\textbf{Normal regression on ModelNet40 by fine-tuning from pre-trained backbones.}}
	\begin{tabular}{ l c| l c}
		\toprule[1.0pt]
		Method & NRErr &Method & NRErr \\
		\hline
		{[P]-Random} & 0.176 & {[D]-Random} & 0.102\\
		{[P]-OcCo~\cite{wang2021unsupervised}} & 0.155 & {[D]-OcCo~\cite{wang2021unsupervised}} & 0.084 \\
		{[P]-CrossPoint~\cite{afham2022crosspoint}} & 0.169 & {[D]-CrossPoint~\cite{afham2022crosspoint}} & 0.098\\
		\rowcolor{light-gray}
		{[P]-PointVST} & \textbf{0.132} & {[D]-PointVST} & \textbf{0.076}\\
		\bottomrule[1.0pt]
	\end{tabular}
	\label{tab:ft-est--modelnet40}
\end{table}

\begin{figure}[t]
	\centering
	\subfigure[]{
		\includegraphics[width=0.48\linewidth]{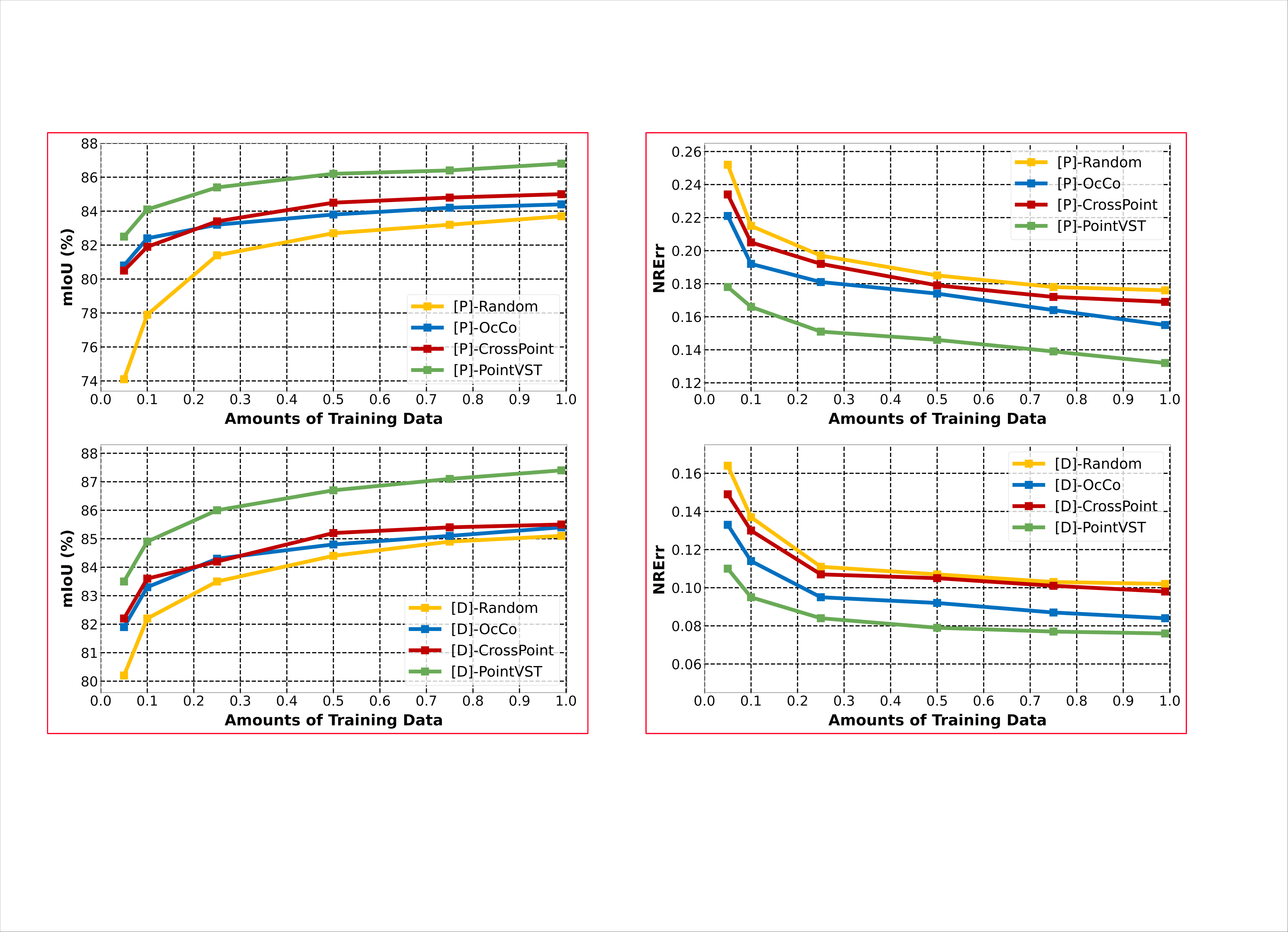}}
	\subfigure[]{
		\includegraphics[width=0.48\linewidth]{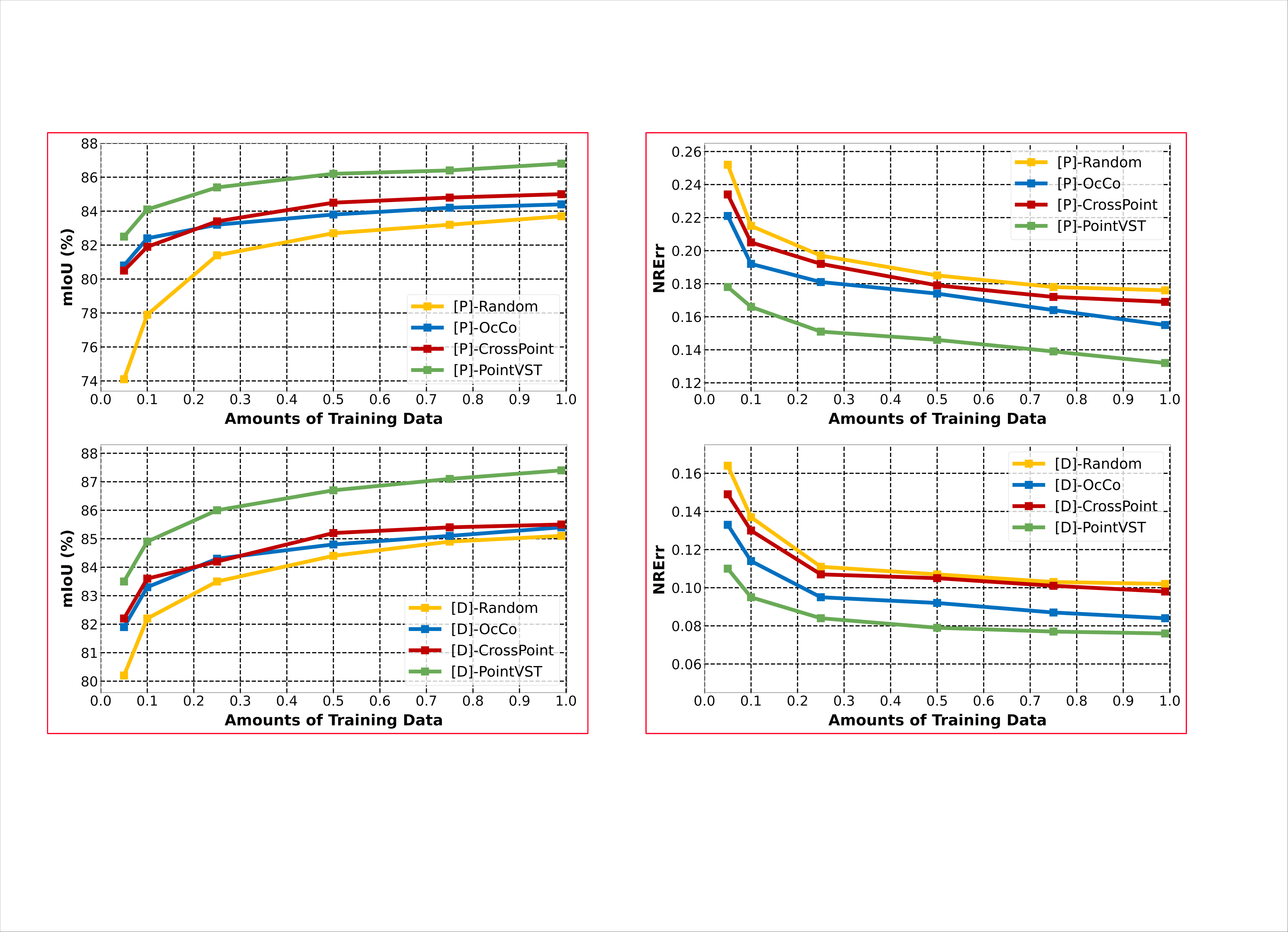}} 
	\caption{\textbf{(a) Semi-supervised part segmentation on ShapeNetPart. (b) Semi-supervised normal estimation on ModelNet40.} We perform fine-tuning from pre-trained backbones using limited training data ($5\%$, $10\%$, $25\%$, $50\%$, $75\%$). The same partial training set is used for all the competing methods for fair comparisons.}
	\label{fig:ps-and-ne-curve}
\end{figure}

Inheriting from previous closely-related approaches \cite{wang2021unsupervised,afham2022crosspoint}, two representative deep set architectures, PointNet~\cite{qi2017pointnet} and DGCNN~\cite{wang2019dynamic} (abbreviated as {[P]} and {[D]}), are adopted as the target point cloud backbones. Throughout our experiments, we involved multiple evaluation protocols, including linear (SVM) probing, fine-tuning, and semi-supervised learning with different amounts of partial training data.

In the following (from Sec.~\ref{sec:exp-obj-cls} to Sec.~\ref{sec:exp-sce-seg}), we mainly focus on comparing the proposed PointVST with influential unsupervised learning and self-supervised pre-training approaches. We especially made extensive comparisons with OcCo~\cite{wang2021unsupervised} and CrossPoint~\cite{afham2022crosspoint}, the two state-of-the-art and most relevant studies that respectively represent generative and contrastive pre-training paradigms. Besides, in Sec.~\ref{sec:ablation-study} we also particularly included comparisons with the very recent MAE-like approaches \cite{pang2022masked,liu2022masked,zhang2022pointmae}, which are specialized for transformer-style backbones and thus not applicable to generic types of deep set architectures \cite{qi2017pointnet,wang2019dynamic}.

\subsection{Object Classification} \label{sec:exp-obj-cls}

ModelNet40~\cite{wu20153d} is a commonly-used 3D shape classification benchmark dataset totally consisting of 12311 object models (9843 for training and 2468 for testing) covering 40 semantic categories. In our experiments, each input point cloud is uniformly composed of 1024 points without auxiliary attributes. Given a pre-trained backbone, we froze its model parameters, and then directly applied it to each point cloud within both the training and testing sets to extract the corresponding collections of vectorized global codewords, based on which a linear SVM is trained and tested.

\begin{table}[t]
	\centering	
	\renewcommand\arraystretch{1.05}
	\setlength{\tabcolsep}{8.0pt}
	\caption{\textbf{Scene-level semantic segmentation on S3DIS under 6-fold cross-validation.}}
	\begin{tabular}{ l c | l c}
		\toprule[1.0pt]
		Method & mIoU & Method & mIoU \\
		\hline		
		{[P]-Random} & 47.6 & {[D]-Random} & 56.1 \\
		{[P]-OcCo~\cite{wang2021unsupervised}} & 54.5 & {[D]-OcCo~\cite{wang2021unsupervised}} & 58.3 \\
		{[P]-CrossPoint~\cite{afham2022crosspoint}} & 54.2 &{[D]-CrossPoint~\cite{afham2022crosspoint}} & 57.9 \\
		\rowcolor{light-gray} 
		{[P]-PointVST} & \textbf{55.7} & {[D]-PointVST} & \textbf{60.1} \\
		\bottomrule[1.0pt]
	\end{tabular}
	\label{tab:ft-sem-seg--s3dis}
\end{table}

\begin{table}[t]
	\scriptsize
	\centering	
	\renewcommand\arraystretch{1.05}
	\setlength{\tabcolsep}{12.0pt}
	\caption{\textbf{S3DIS semantic segmentation results in Area 5 by fine-tuning from the pre-trained PointNext-XL backbone.}}
	\begin{tabular}{ c | c }
		\toprule[1.0pt]
		PointNext-XL~\cite{qian2022pointnext} & PointNext-XL~\cite{qian2022pointnext} \textit{w/} PointVST \\
		\hline
		70.5 & \textbf{71.8} \\
		\bottomrule[1.0pt]
	\end{tabular}
	\label{ltab:pointnext-backbone-s3dis-area-5}
\end{table}

Table~\ref{tab:svm-cls--modelnet40} compares the linear classification performances of different unsupervised and self-supervised methods under the measurement of overall accuracy (OAcc), in which our PointVST outperforms all the competing methods with large margins. In particular, compared with the second best method CrossPoint that also shares cross-modal characteristics, we still demonstrate prominent performance superiority, i.e., $1.1\%$ and $0.9\%$ accuracy improvements for PointNet and DGCNN backbones, respectively. Besides, we further explored the rotation robustness of different methods, which can better reveal the representation capability of the learned geometric features. Accordingly, all the competing methods are equipped with random rotation for data augmentation purposes during pre-training. Quantitative results are presented in Table~\ref{tab:svm-cls-rot--modelnet40}. For the z/z setting, all the competing methods show relatively satisfactory performances, and our PointVST still takes the leading position. For the much more challenging SO3/SO3 setting, though all methods suffer from significant degradation, our PointVST shows larger performance gains.

In addition to transferring features exported from frozen backbone encoders, we also attempted to use the pre-trained model parameters for backbone network initialization, after which the whole learning framework is fine-tuned in a task-specific manner. Here, we experimented with ScanObjectNN~\cite{uy2019revisiting} (the OBJ-BG split), a much more challenging real-scanned object classification benchmark. Point clouds in this dataset are typically noisy, incomplete, non-uniform, and interfered by background context. As illustrated in Table~\ref{tab:ft-cls--scanobjectnn}, our PointVST achieves the best performances with both backbones, effectively demonstrating our transferability from synthetic models to real-world point cloud scans.

\subsection{Part Segmentation} \label{sec:exp-par-seg}

ShapeNetPart~\cite{yi2016scalable} is a popular large-scale 3D object segmentation dataset with 50 classes of part-level annotations coming from 16 different object-level categories. Under the official split, there are 14007 models for training and 2874 models for testing. Each input point cloud uniformly contains 2048 points without auxiliary attributes.

In contrast to the preceding task scenario (shape classification) of global geometry recognition, here the point-wise semantic prediction task requires to extract much more fine-grained and discriminative feature descriptions.

\begin{table}[t]
	\centering	
	\renewcommand\arraystretch{1.05}
	\setlength{\tabcolsep}{10.0pt}
	\caption{\textbf{Effects of pre-training PointVST with different image translation objectives under ModelNet40 linear SVM classification.} Within brackets, we emphasize the performance degradation of each variant relative to the full implementation.}
	\begin{tabular}{ c c c | c c }
		\toprule[1.0pt]
		$\mathcal{L}_d$ & $\mathcal{L}_s$ & $\mathcal{L}_c$ & [P]-PointVST & [D]-PointVST \\
		\hline		
		\ding{55} &  &  & 89.3 ($-$0.9) & 91.4 ($-$0.7) \\
		& \ding{55} &  & 89.7 ($-$0.5) & 91.5 ($-$0.6) \\
		&  & \ding{55} & 89.9 ($-$0.3) & 91.9 ($-$0.2) \\
		\bottomrule[1.0pt]
	\end{tabular}
	\label{tab:ablation--trans-obj}
\end{table}

\begin{table}[t]
	\centering	
	\renewcommand\arraystretch{1.05}
	\setlength{\tabcolsep}{8.0pt}
	\caption{\textbf{Comparisons of pre-training PointVST with ground-truth images rendered from meshes (M-Rendered) and points (P-Rendered) under ModelNet40 linear SVM classification.}}
	\begin{tabular}{ c c | c c }
		\toprule[1.0pt]
		\multicolumn{2}{c|}{[P]-PointVST}  & \multicolumn{2}{c}{[D]-PointVST}  \\ 
		\hline
		\multicolumn{1}{c|}{M-Rendered} & P-Rendered & \multicolumn{1}{c|}{M-Rendered} & P-Rendered \\ 
		\hline
		\multicolumn{1}{c|}{90.2} & 90.0 & \multicolumn{1}{c|}{92.1} & 91.8 \\
		\bottomrule[1.0pt]
	\end{tabular}
	\label{tab:ablation--render}
\end{table}

Given a pre-trained backbone, we performed fine-tuning for the whole segmentation framework. Table~\ref{tab:ft-seg--shapenetpart} reports the performances of different approaches under the measurement of mean interaction-over-union (mIoU). It is observed that our PointVST significantly outperforms other methods. Meanwhile, we noticed that the performance improvements relative to their DGCNN baselines (with random initialization) brought by OcCo and CrossPoint are relatively limited (with only $0.3\%$ and $0.4\%$ gains, respectively), while ours can reach $2.3\%$. Even for the less expressive PointNet backbone, its fine-tuning result reaches $86.8\%$, which turns to be highly competitive. As for PointContrast, we can observe that its relative performance gain is much smaller than ours. Besides, we further explored fine-tuning using different amounts of labeled training data. As illustrated in Figure~\ref{fig:ps-and-ne-curve}, our PointVST consistently shows significant performance gains over all the competing methods.

\subsection{Normal Estimation} \label{sec:exp-nor-est}

Previous self-supervised pre-training works focus on the evaluation of high-level semantic understanding tasks such as classification and segmentation, but ignore verifications on low-level geometry processing tasks that are also highly valuable in various applications. To fill in this gap, we further performed normal estimation by fine-tuning from pre-trained backbones on ModelNet40, where each input point cloud contains 1024 points and the corresponding ground-truth normal vectors can be easily deduced from mesh faces. Note that for approaches \cite{guerrero2018pcpnet} specialized for the normal estimation task, the network consumes as inputs local patches cropped from dense point clouds, while in our setting the inputs are complete point cloud objects with much sparser density. In practice, considering that the computed ground-truth normals usually suffer from flipped directions, we choose to perform unoriented, instead of oriented \cite{li2023shs,xu2023globally}, normal estimation. We adopted the absolute cosine distance (proposed in \cite{qi2017pointnet}) as the optimization objective as well as evaluation metric, which can be formulated as
\begin{equation}
	\mathrm{NRErr} = 1 - \texttt{abs}( \frac{ \mathbf{n}_\mathrm{gt} \cdot \mathbf{n}_\mathrm{pr} }{ \lVert \mathbf{n}_\mathrm{gt} \rVert_2 \cdot \lVert \mathbf{n}_\mathrm{pr} \rVert_2  } ),
\end{equation}
where $\mathbf{n}_\mathrm{pr}$ and $\mathbf{n}_\mathrm{gt}$ denote the regressed and ground-truth normal vectors, respectively.

Quantitative results are presented in Table~\ref{tab:ft-est--modelnet40}, where an interesting observation is that OcCo outperforms CrossPoint for both backbones, and the performance gains brought by CrossPoint become even negligible. This phenomenon may indicate that a straightforward contrastive learning strategy implemented across heterogeneous 2D image and 3D point cloud features can barely benefit low-level fine-grained geometry processing. Besides, as for fine-tuning with different amounts of labeled training data, as shown in Figure~\ref{fig:ps-and-ne-curve}, CrossPoint still shows inferior performances against OcCo, while our PointVST consistently takes the leading position with obvious margins.

\begin{figure}[t]
	\centering
	\includegraphics[width=1.0\linewidth]{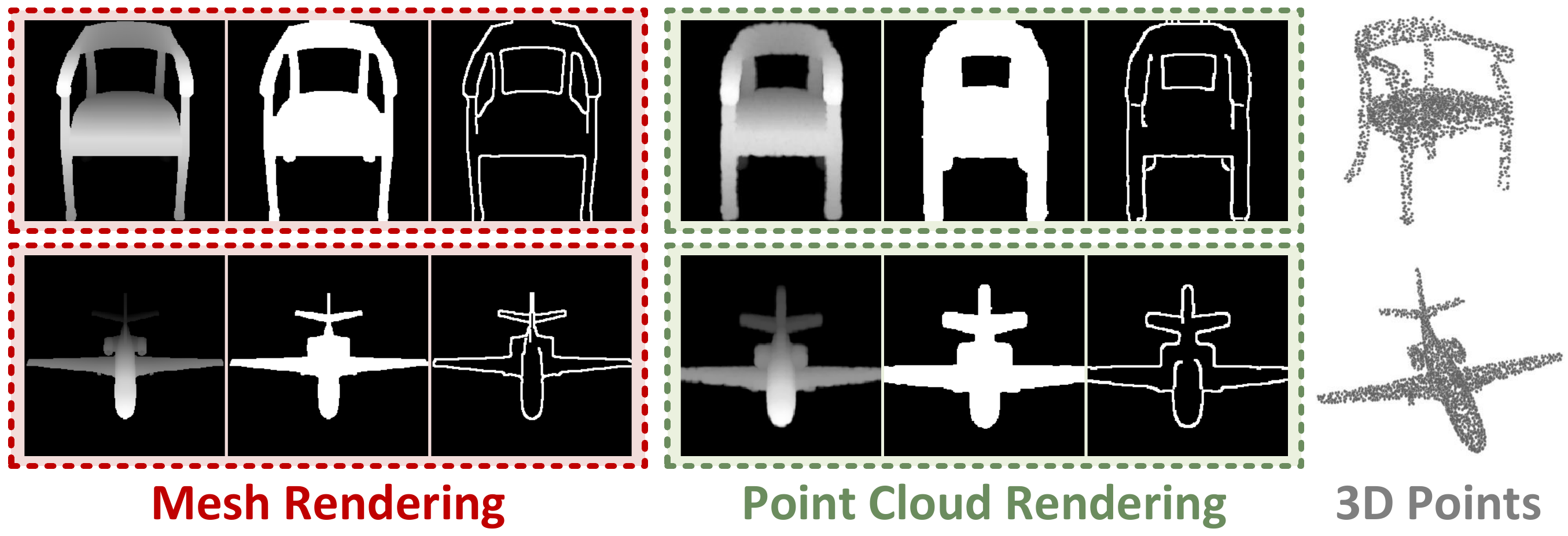}
	\caption{\textbf{Visual comparisons of ground-truth images rendered from meshes and sparse point clouds.}}
	\label{fig:different-rendering-schemes}
\end{figure}

\subsection{Scene Segmentation} \label{sec:exp-sce-seg}

In addition to object-centric tasks, we also experimented with indoor scene segmentation on S3DIS~\cite{armeni20163d}, a commonly-used benchmark dataset, containing $271$ single rooms from $6$ different areas, with over $270$ million densely annotated points covering $13$ semantic categories. The pre-processing procedures follow the original baselines \cite{qi2017pointnet,wang2019dynamic}. 

Following the evaluation protocol of previous work \cite{wang2021unsupervised}, here we aim to verify the transferability of the pre-trained backbones from the source object-level domain to the target scene-level domain. As shown in Table~\ref{tab:ft-sem-seg--s3dis}, all the competing methods can bring different degrees of performance boosts relative to randomly initialized baselines, and our PointVST also shows obvious performance superiority. Besides, we also made additional efforts to apply our PointVST to recent state-of-the-art backbone \cite{qian2022pointnext} to explore the potential of our pre-training effects. As compared in Table~\ref{ltab:pointnext-backbone-s3dis-area-5}, the pre-training process still brings obvious performance gains.

\subsection{Ablation Study} \label{sec:ablation-study}

\noindent \textbf{\textit{Image translation objectives}.} As described in Eq.~(\ref{eqn-loss-overall}), in the full implementation, our translation objectives involve three different forms of 2D rendered images, i.e., depth, contour, and silhouette maps. To reveal their necessity and influence, we conducted experiments under the setting of ModelNet40 linear SVM classification by respectively removing each of the three loss terms. As indicated in Table~\ref{tab:ablation--trans-obj}, all the three image supervisions turn to be necessary. The removal of $\mathcal{L}_d$ leads to the most severe performance degradation, because depth maps convey the richest geometric information.
\vspace{0.25cm}

\noindent \textbf{\textit{Image rendering strategies}.} Different ways of ground-truth image creation can influence the actual pre-training effects. Although mesh rendering is chosen in our implementation, we demonstrate that point cloud rendering also works well. Here, for the setting of ModelNet40 SVM classification, we adopted Pytorch3D \cite{ravi2020pytorch3d} library's rasterizer to render images directly from point clouds in the pre-training phase. Typical visual examples are presented in Figure~\ref{fig:different-rendering-schemes}. As reported in Table~\ref{tab:ablation--render}, compared with mesh rendering (with $90.2$\% and $92.1$\% OAcc for backbones [P] and [D]), switching to point cloud rendering causes slight performance degradation, but still turns to be highly effective.
\vspace{0.25cm}

\begin{table*}[t]
	\renewcommand\arraystretch{1.05}
	\centering
	\caption{\textbf{Domain transfer modes involved in previous influential works on self-supervised 3D point cloud representation learning.} The left and the right side of ``$\Rightarrow$'' respectively denotes the source data domain (for pre-training) and the target data domain (for fine-tuning). For simplicity, we use the notations ``$\mathcal{O}$'' and ``$\mathcal{S}$'' to respectively represent the \textbf{object-level} (e.g., ShapeNet~\cite{chang2015shapenet}, ModelNet~\cite{wu20153d}) and \textbf{scene-level} (e.g., ScanNet~\cite{dai2017scannet}, S3DIS~\cite{armeni20163d}) point cloud data domains. In the last column, we particularly denote whether the corresponding work considers the data domain gap between object-level and scene-level point clouds.}
	\setlength{\tabcolsep}{5.0mm}{
		\begin{tabular}{ l | c c c c | c }
			\toprule[1.0pt]
			\multirow{2}{*}{Previous Works} & \multicolumn{4}{c|}{Source-to-Target Transfer Modes} & \multirow{2}{*}{With Gap Across $\mathcal{O}$ and $\mathcal{S}$} \\
			\cline{2-5}
			& \multicolumn{1}{c|}{$\mathcal{O}$ $\Rightarrow$ $\mathcal{O}$} & \multicolumn{1}{c|}{$\mathcal{S}$ $\Rightarrow$ $\mathcal{S}$} & \multicolumn{1}{c|}{$\mathcal{O}$ $\Rightarrow$ $\mathcal{S}$} & $\mathcal{S}$ $\Rightarrow$ $\mathcal{O}$ & \\
			\hline
			PointContrast~\cite{xie2020pointcontrast} & \multicolumn{1}{c|}{} & \multicolumn{1}{c|}{\ding{51}} & \multicolumn{1}{c|}{} & \ding{51} & \ding{51} \\
			DepthContrast~\cite{zhang2021self} & \multicolumn{1}{c|}{} & \multicolumn{1}{c|}{\ding{51}} & \multicolumn{1}{c|}{} & \ding{51} & \ding{51} \\
			OcCo~\cite{wang2021unsupervised} & \multicolumn{1}{c|}{\ding{51}} & \multicolumn{1}{c|}{} & \multicolumn{1}{c|}{\ding{51}} &  & \ding{51} \\
			STRL~\cite{huang2021spatio} & \multicolumn{1}{c|}{\ding{51}} & \multicolumn{1}{c|}{\ding{51}} & \multicolumn{1}{c|}{\ding{51}} & \ding{51} & \ding{51} \\
			Point-BERT~\cite{yu2022point} & \multicolumn{1}{c|}{\ding{51}} & \multicolumn{1}{c|}{} & \multicolumn{1}{c|}{} &  & \ding{55} \\
			CrossPoint~\cite{afham2022crosspoint} & \multicolumn{1}{c|}{\ding{51}} & \multicolumn{1}{c|}{} & \multicolumn{1}{c|}{} &  & \ding{55} \\
			Point-MAE~\cite{pang2022masked} & \multicolumn{1}{c|}{\ding{51}} & \multicolumn{1}{c|}{} & \multicolumn{1}{c|}{} &  & \ding{55} \\
			MaskPoint~\cite{liu2022masked} & \multicolumn{1}{c|}{\ding{51}} & \multicolumn{1}{c|}{\ding{51}} & \multicolumn{1}{c|}{} &  & \ding{55} \\
			Point-M2AE~\cite{zhang2022pointmae} & \multicolumn{1}{c|}{\ding{51}} & \multicolumn{1}{c|}{\ding{51}} & \multicolumn{1}{c|}{} &  & \ding{55} \\
			\hline
			\rowcolor{light-gray}
			PointVST~(Ours) & \multicolumn{1}{c|}{\ding{51}} & \multicolumn{1}{c|}{\ding{51}} & \multicolumn{1}{c|}{\ding{51}} & \ding{51} & \ding{51} \\
			\bottomrule[1.0pt]
	\end{tabular}}
	\label{tab:summary}
\end{table*}

\begin{table}[t]
	\centering	
	\renewcommand\arraystretch{1.05}
	\setlength{\tabcolsep}{10.0pt}
	\caption{\textbf{Performances of {[D]-PointVST} with different source (to be pre-trained) and target (to be fine-tuned) data domains. $\mathcal{O}$ and $\mathcal{S}$ denote object-level and scene-level, respectively.}}
	\begin{tabular}{ c c c c }
		\toprule[1.0pt]
		Source Data & Target Data  & Transfer Mode & mIoU \\
		\hline
		ShapeNet & S3DIS & $\mathcal{O}$ $\Rightarrow$ $\mathcal{S}$ & 60.1 \\
		ScanNet & S3DIS & $\mathcal{S}$ $\Rightarrow$ $\mathcal{S}$ & 63.7 \\
		ShapeNet & ShapeNetPart & $\mathcal{O}$ $\Rightarrow$ $\mathcal{O}$ & 87.4 \\
		ScanNet & ShapeNetPart & $\mathcal{S}$ $\Rightarrow$ $\mathcal{O}$  & 86.5 \\
		\bottomrule[1.0pt]
	\end{tabular}
	\label{tab:ablation--cross}
\end{table}

\noindent \textbf{\textit{Pre-training data domains}.} Transferability is a critical factor when evaluating the superiority of self-supervised representation learning approaches. Depending on different data characteristics, common point cloud processing tasks can be classified as object-level and scene-level. Accordingly, in different previous works, the source point cloud data domains selected for implementing the pre-training process can be object-level \cite{wang2021unsupervised,yu2022point,afham2022crosspoint,pang2022masked}, scene-level \cite{xie2020pointcontrast,zhang2021self}, or both \cite{huang2021spatio,liu2022masked}. To explore the scalability of PointVST for pre-training with scene-level point clouds, here we adopted the popular ScanNet~\cite{dai2017scannet} dataset to prepare the corresponding pre-training data, where the 2D ground-truth images are directly rendered from point clouds of scene blocks/patches. For data preparation on ScanNet~\cite{dai2017scannet}, we cropped sub-regions from the whole point cloud scene (with grid-subsampling~\cite{hu2020randla}) using both kNN-based and area-based strategies, such that each cropped point cloud block/patch contains $4096$ 3D spatial points (without auxiliary attributes). As summarized in Table~\ref{tab:summary}, compared with most previous influential works on self-supervised point cloud representation learning, in this paper, we explored more comprehensive transfer modes between object-level and scene-level point cloud data domains, which can better validate the effectiveness and generalizability of our proposed PointVST.

We report fine-tuning (with backbone [D]) performances in Table~\ref{tab:ablation--cross}, where the preceding results with ShapeNet pre-training are also pasted (in the first and third rows) for convenience. It is observed that PointVST shows satisfactory improvements under different transfer modes. And we also noticed that better results can be achieved given a smaller source-target domain gap (e.g., $\mathcal{O}$ $\Rightarrow$ $\mathcal{O}$, $\mathcal{S}$ $\Rightarrow$ $\mathcal{S}$).
\vspace{0.25cm}

\begin{table}[h]
	\centering	
	\renewcommand\arraystretch{1.05}
	\setlength{\tabcolsep}{20.0pt}
	\caption{\textbf{Outdoor scene segmentation on Toronto3D by fine-tuning from the DGCNN backbone pre-trained on ShapeNet.}}
	\begin{tabular}{ l c }
		\toprule[1.0pt]
		Method & mIoU \\
		\hline
		{[D]-Random} & 62.3 \\
		{[D]-OcCo~\cite{wang2021unsupervised}} & 63.6 \\
		{[D]-CrossPoint~\cite{afham2022crosspoint}} & 63.1 \\
		\rowcolor{light-gray}
		{[D]-PointVST} & \textbf{64.9} \\
		\bottomrule[1.0pt]
	\end{tabular}
	\label{tab:toronto-comparison}
\end{table}

\begin{table}[h]
	\centering	
	\renewcommand\arraystretch{1.05}
	\setlength{\tabcolsep}{12.0pt}
	\caption{\textbf{Comparisons of Toronto3D segmentation performances with different source domains for backbone pre-training, including two object datasets (ShapeNet and ScanObjectNN) and one scene dataset (ScanNet).}}
	\begin{tabular}{ l l c }
		\toprule[1.0pt]
		Method & Source Domain & mIoU \\
		\hline
		{[D]-PointVST} & ShapeNet~\cite{chang2015shapenet} ($\mathcal{O}$) & 64.9 \\
		{[D]-PointVST} & ScanObjectNN~\cite{uy2019revisiting} ($\mathcal{O}$) & 64.5 \\
		{[D]-PointVST} & ScanNet~\cite{dai2017scannet} ($\mathcal{S}$) & 65.7 \\
		\bottomrule[1.0pt]
	\end{tabular}
	\label{tab:toronto-diff-source}
\end{table}

\noindent \textbf{\textit{Effectiveness on outdoor LiDAR data}.} Toronto3D~\cite{tan2020toronto} is a large-scale LiDAR point cloud dataset for benchmarking real-scene semantic segmentation, which consists of 78.3 million densely-annotated points covering approximately 1KM of outdoor urban space. There are totally $8$ different categories of semantic labels.

Following the official split \cite{tan2020toronto}, the section L002 is used for testing and the rest three sections (L001, L003, L004) are used for training.  The raw data pre-processing procedures follow \cite{hu2021learning}. Particularly, as analyzed in \cite{hu2021learning}, it is basically impossible to differentiate some classes (e.g., \textit{road marking} and \textit{road}) given only 3D coordinates of points. Hence, as also explored in \cite{hu2021learning}, we further included RGB color attributes as additional input information (for all the competing methods) to make the corresponding evaluations more discriminative. Table~\ref{tab:toronto-comparison} reports quantitative performances of different approaches, where we experimented with the DGCNN backbone pre-trained on ShapeNet. Our PointVST achieves $2.6\%$ improvement over the baseline DGCNN and outperforms both OcCo and CrossPoint. Besides, as shown in Table~\ref{tab:toronto-diff-source}, we also explored the effects of pre-training with different domains of source datasets. We can observe that the downstream performance is still competitive even with a much smaller pre-training dataset (i.e., ScanObjectNN). Switching to ScanNet for pre-training leads to better performances since the domain map between source and target data becomes smaller.
\vspace{0.25cm}

\begin{table}[t]
	\centering	
	\renewcommand\arraystretch{1.05}
	\setlength{\tabcolsep}{10.0pt}
	\caption{\textbf{Effectiveness verifications of AVS-Pool under ModelNet40 linear SVM classification}, where Avg-Pool, Max-Pool, Att-Pool, and AVS-Pool-Uns respectively denote average-pooling, max-pooling, attentive pooling \cite{yang2020robust}, and AVS-Pool without explicitly supervising $\mathbf{S_v}$ (i.e., removing $\mathcal{C}_v$ from $\mathcal{L}_\mathrm{overall}$). GT-Vis means replacing adaptive visibility mask prediction with ground-truth visibility. Within brackets, we emphasize the performance degradation of each variant relative to the full implementation.}
	\begin{tabular}{ c | c c }
		\toprule[1.0pt]
		Strategy & [P]-PointVST & [D]-PointVST \\
		\hline		
		Avg-Pool & 88.8 ($-$1.4) & 90.5 ($-$1.6) \\
		Max-Pool & 89.3 ($-$0.9) & 91.1 ($-$1.0) \\
		Att-Pool~\cite{yang2020robust} & 89.6 ($-$0.6) & 90.9 ($-$1.2) \\
		AVS-Pool-Uns & 89.7 ($-$0.5) & 91.4 ($-$0.7) \\
		GT-Vis & 89.9 ($-$0.3) & 91.6 ($-$0.5) \\
		\bottomrule[1.0pt]
	\end{tabular}
	\label{tab:ablation--avs-pool}
\end{table}

\begin{table}[t]
	\centering	
	\renewcommand\arraystretch{1.05}
	\setlength{\tabcolsep}{5.0pt}
	\caption{\textbf{Comparison with MAE-like pre-training approaches specialized for transformer-style point cloud backbones.}}
	\begin{tabular}{ l c c c }
		\toprule[1.0pt]
		Method & ModelNet40  & ScanObjectNN & ShapeNetPart \\
		\hline
		{[T]-Point-MAE}~\cite{pang2022masked} & 93.8 & 85.2 & 86.1 \\
		{[T]-MaskPoint}~\cite{liu2022masked} & 93.8 & 84.6 & 86.0 \\
		{[Tm]-Point-M2AE}~\cite{zhang2022pointmae} & 94.0 & 86.4 & 86.5 \\
		\hline
		\rowcolor{light-gray} 
		{[T]-PointVST} & \textbf{94.1} & \textbf{86.6} & \textbf{87.6} \\
		\bottomrule[1.0pt]
	\end{tabular}
	\label{tab:ablation--transformer}
\end{table}

\noindent \textbf{\textit{Effectiveness of AVS-Pool}.} In our technical implementation, deducing a view-specific global codeword from view-conditioned point-wise embeddings relies on the proposed AVS-Pool. To evaluate its necessity, we designed multiple variants of PointVST by replacing AVS-Pool with different pooling strategies. In addition to common average-pooling and max-pooling, we also introduced a more advanced attentive aggregation technique \cite{yang2020robust} that has been integrated in various low-level and high-level point cloud processing tasks \cite{hu2020randla,hu2021learning,wang2022rangeudf}. As shown in Table~\ref{tab:ablation--avs-pool}, average-pooling suffers from the most severe performance degradation due to its lack of selectivity, max-pooling and attentive pooling show better performances since they are more suitable for view-specific aggregation. The comparison between AVS-Pool-Uns and the full implementation can validate the necessity of our proposed visibility constraint $\mathcal{C}_v$. Moreover, as indicated in the last row for the GT-Vis variant, although directly using ground-truth visibility seems more straightforward, it tends to be sub-optimal compared with adaptively learning the visibility mask. We reason that the adaptive prediction process itself can guide the model to better extract view-specific patterns.
\vspace{0.25cm}

\begin{table}[t]
	\centering	
	\renewcommand\arraystretch{1.05}
	\setlength{\tabcolsep}{12.0pt}
	\caption{\textbf{Influence of position encoding (P.E.) for view representation on ModelNet40 linear SVM classification.}}
	\begin{tabular}{ c | c c }
		\toprule[1.0pt]
		View Representation & [P]-PointVST & [D]-PointVST \\
		\hline
		w/ P.E. & \textbf{90.2} & \textbf{92.1} \\
		w/o P.E. & 89.8 & 91.9 \\
		\bottomrule[1.0pt]
	\end{tabular}
	\label{tab:ablation--p.e.}
\end{table}

\begin{table}[t]
	\centering	
	\renewcommand\arraystretch{1.05}
	\setlength{\tabcolsep}{12.0pt}
	\caption{\textbf{Influence of the length of view-specific global codeword $\mathbf{g_v}$ on ModelNet40 linear SVM classification.}}
	\begin{tabular}{ c | c c }
		\toprule[1.0pt]
		Codeword Length & [P]-PointVST & [D]-PointVST \\
		\hline
		2048 & 90.2 & \textbf{92.1} \\
		1024 & 89.7 & 91.8 \\
		4096 & \textbf{90.3} & 91.7 \\
		\bottomrule[1.0pt]
	\end{tabular}
	\label{tab:ablation--len-gv}
\end{table}

\noindent \textbf{\textit{Transformer-oriented pre-training}.} As mentioned earlier, MAE-like pre-training frameworks \cite{yu2022point,pang2022masked,liu2022masked,zhang2022pointmae} that are specialized for transformer-style 3D point cloud backbones also attract much attention for their impressive performances, despite the inapplicability to many other widely-used (non-transformer) deep set architectures.

To fairly form apples-to-apples comparisons, we applied PointVST to pre-train a standard transformer backbone \cite{pang2022masked,liu2022masked} (abbreviated as {[T]}), and then performed fine-tuning for object classification on ModelNet40 and ScanObjectNN (the PB-T50-RS split), and part segmentation on ShapeNetPart. Note that Point-M2AE~\cite{zhang2022pointmae} is built upon a more advanced multi-scale transformer (denoted as {[Tm]}) backbone. Following these works, we used test voting \cite{liu2019relation} on ModelNet40, but not for ScanObjectNN and ShapeNetPart. As illustrated in Table~\ref{tab:ablation--transformer}, our PointVST still shows leading performances when applied to the transformer-style backbones, which are known to have greater model capacity.
\vspace{0.25cm}

\noindent \textbf{\textit{Positional Encoding of Raw Camera Positions}.} As introduced in Eq.~(\textcolor{red}{1}), the vectorized viewpoint indicator $\mathbf{v}$ is obtained via simple positional encoding. Here, we removed the two non-linear transformation layers ($\mathcal{F}_{\phi}$, $\mathcal{F}_{\lambda}$) and directly used raw latitude and longitude angles $(\phi_{\mathrm{lat}},\lambda_{\mathrm{lon}})$ for the subsequent processing procedures.

As reported in Table~\ref{tab:ablation--p.e.}, compared with the full implementation, the removal of positional encoding respectively leads to $0.4\%$ and $0.2\%$ performance degradation for our [P]-PointVST and [D]-PointVST on ModelNet40 linear SVM classification.
\vspace{0.25cm}

\noindent \textbf{\textit{Length of View-Specific Global Codeword}.} In all our experiments, we uniformly configured the length of the produced view-specific global codeword $\mathbf{g_v}$ as $2048$. Here, we further experimented with $1024$ and $4096$ to explore its influence. 

As reported in Table~\ref{tab:ablation--len-gv}, reducing the length to $1024$ leads to different degrees of performance degradation for both [P]-PointVST and [D]-PointVST. When increasing the length to $4096$, the performance of [D]-PointVST decreases by $0.4\%$, while [P]-PointVST is slightly boosted by $0.1\%$.
\vspace{0.25cm}

\begin{table}[t]
	\centering	
	\renewcommand\arraystretch{1.05}
	\setlength{\tabcolsep}{12.0pt}
	\caption{\textbf{Influence of different dimensions of translated images on ModelNet40 linear SVM classification.}}
	\begin{tabular}{ c | c c }
		\toprule[1.0pt]
		Image Dimension & [P]-PointVST & [D]-PointVST \\
		\hline
		$128 \times 128$ & \textbf{90.2} & \textbf{92.1} \\
		$256 \times 256$ & 90.0 & 92.0 \\
		$64 \times 64$ & 89.8 & 91.6 \\
		\bottomrule[1.0pt]
	\end{tabular}
	\label{tab:ablation--dim-trans-images}
\end{table}

\begin{figure}[t]
	\centering
	\includegraphics[width=1.0\linewidth]{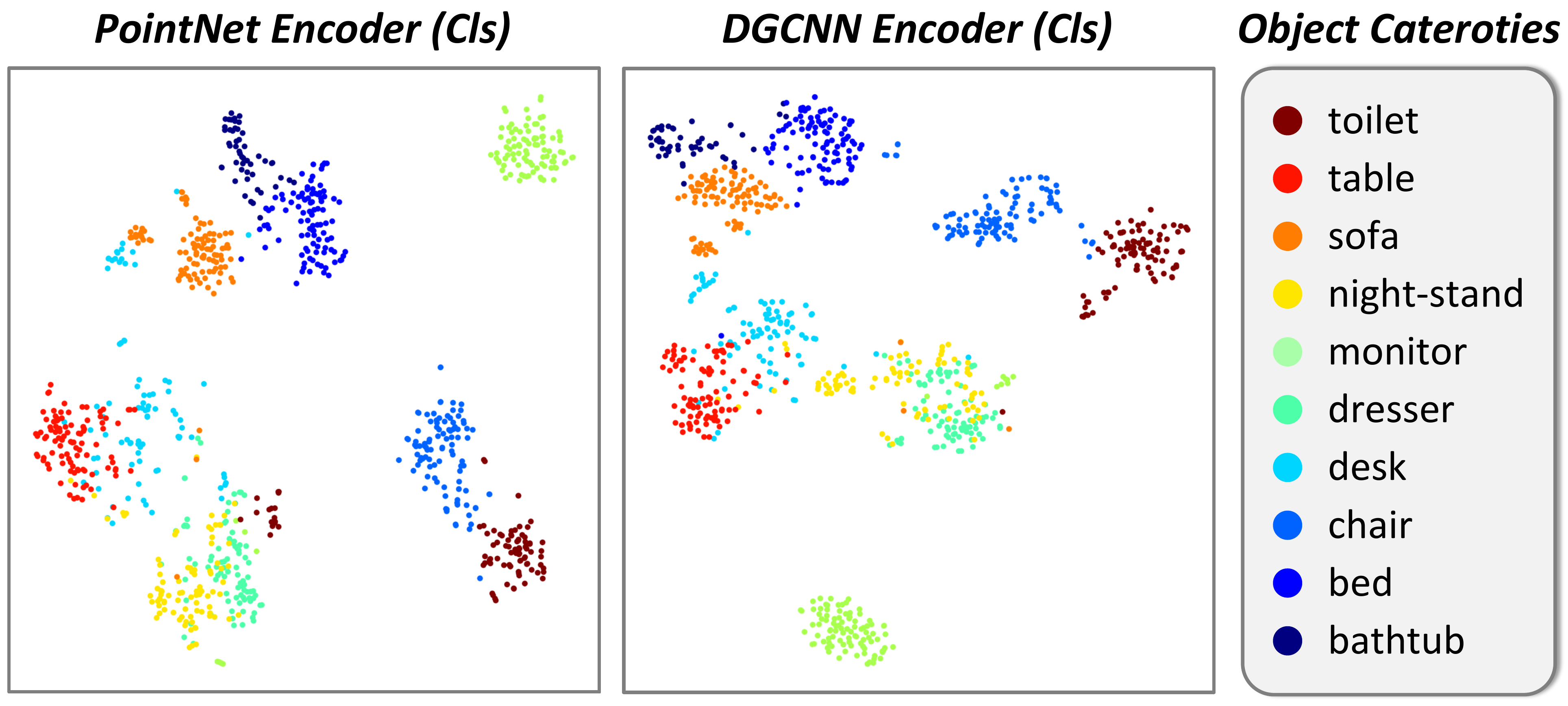}
	\caption{\textbf{Visualization of features exported from our pre-trained PointNet and DGCNN backbones on ModelNet10.}}
	\label{fig:tsne-visualization}
\end{figure}

\noindent \textbf{\textit{Dimension of Translated 2D Images}.} Throughout our experiments, we uniformly configured the dimension of translated images as $128 \times 128$. Here, we also experimented with larger $256 \times 256$ and smaller $64 \times 64$ dimensions.

As compared in Table~\ref{tab:ablation--dim-trans-images}, the performances of our [P]-PointVST and [D]-PointVST respectively decrease by $0.2\%$ and $0.1\%$ when using larger image dimension of $256 \times 256$. This may indicate that the actual pre-training effects can be weakened when the pretext task becomes much more difficult (i.e., recovering more detailed geometric details in the image domain). When the image dimension is set to smaller $64 \times 64$, the performance degradation further reaches $0.4\%$ and $0.5\%$ for [P]-PointVST and [D]-PointVST, respectively.
\vspace{0.25cm}

\noindent \textbf{\textit{Visualization of Feature Distribution}.} As shown in Figure~\ref{fig:tsne-visualization}, we adopted the classic t-SNE~\cite{van2008visualizing} technique for visualizing the distribution of the learned feature representations exported from our pre-trained backbone encoders on the test split of ModelNet10. We can observe that our PointVST can contribute satisfactory intra-class clustering and inter-class separability effects in the pre-trained backbone feature encoders.
\vspace{0.25cm}

\begin{figure}[t]
	\centering
	\includegraphics[width=1.0\linewidth]{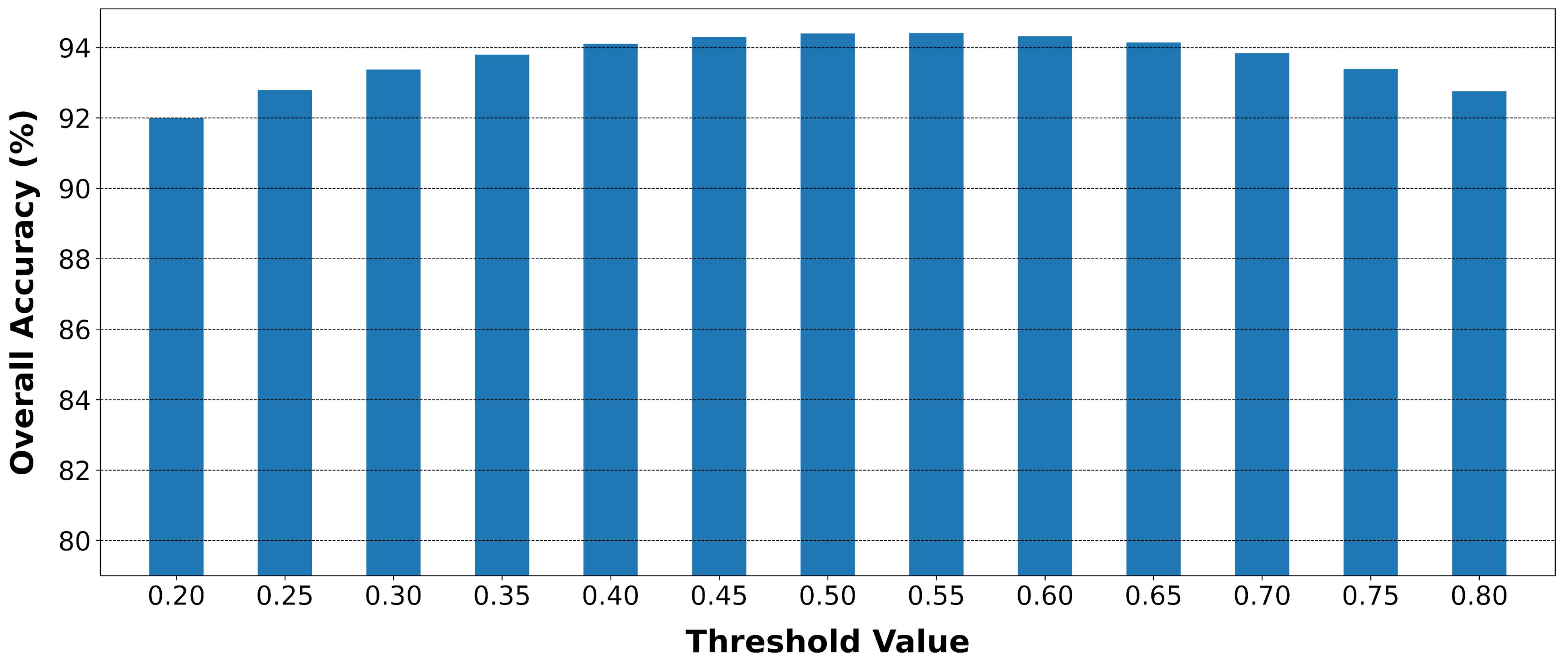}
	\caption{\textbf{Overall accuracy of the predicted point-wise visibility status when pre-training the DGCNN (classification-oriented) backbone encoder.} We applied different thresholding values (from $0.20$ to $0.80$ with the uniform interval of $0.05$) for the binarization of the raw score values in $\mathbf{S_v}$.}
	\label{fig:visibility-accuracy-bar}
\end{figure}

\noindent \textbf{\textit{Pretext Task Performances}.} Under the development protocols of self-supervised pre-training, the \textit{core focus} lies in whether the target backbone encoder is effectively guided to learn transferable features and to what extent the downstream task performance can be boosted. However, to facilitate a more intuitive and comprehensive understanding, we still provided both qualitative and quantitative results of the pretext task performance within our proposed PointVST.

\begin{figure}[h]
	\centering
	\includegraphics[width=1.0\linewidth]{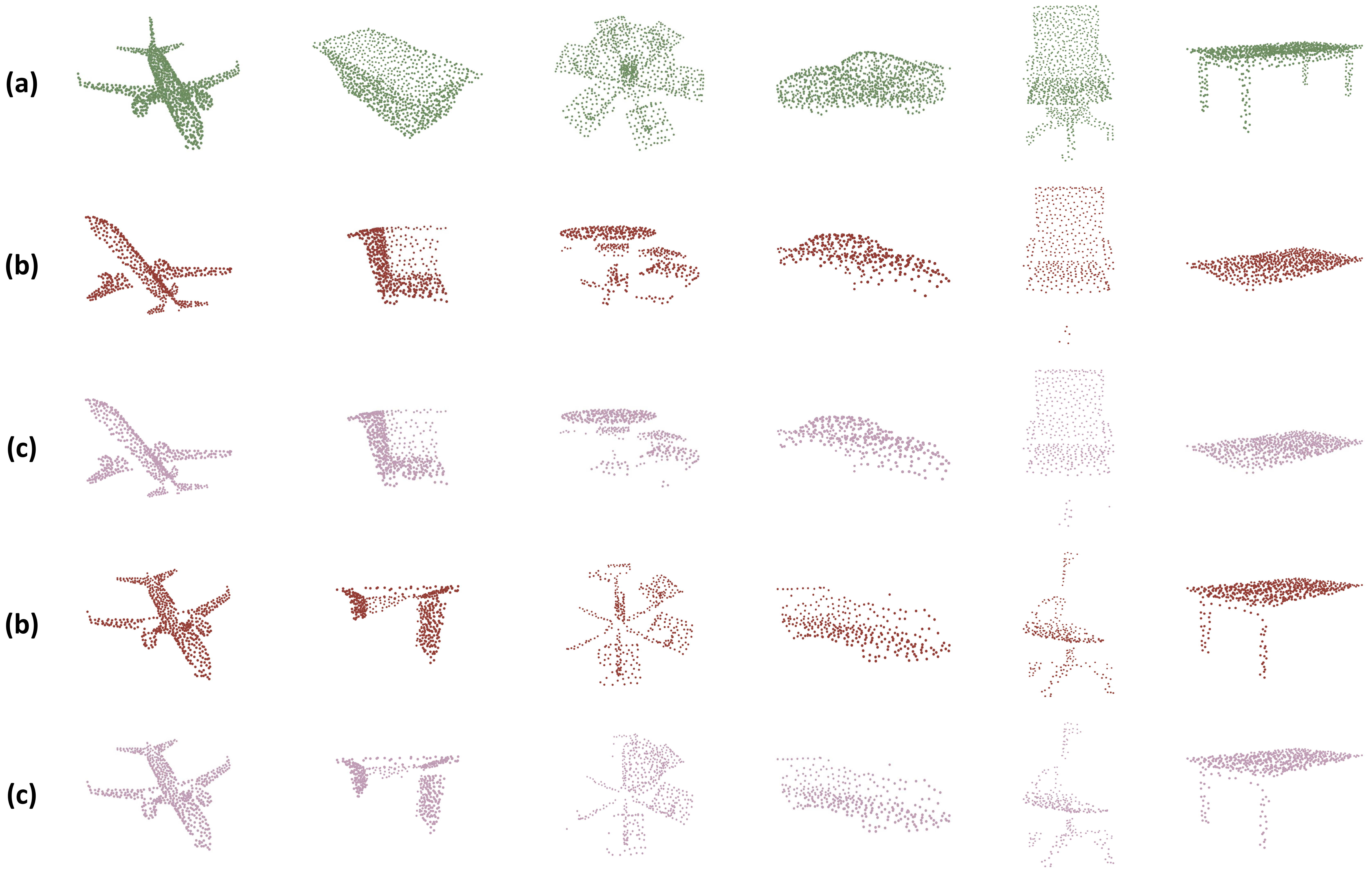}
	\caption{\textbf{Visual examples of point-wise visibility checking,} where we presented (b) ground-truth and (c) our predicted visible points with respect to randomly specified viewpoints by masking the corresponding (a) input point clouds $\mathbf{P}$ with $\mathbf{S^{\prime}_v}$ and $\mathbf{S_v}$.}
	\label{fig:visibility-visual-examples}
\end{figure}

\begin{figure}[t]
	\centering
	\includegraphics[width=0.98\linewidth]{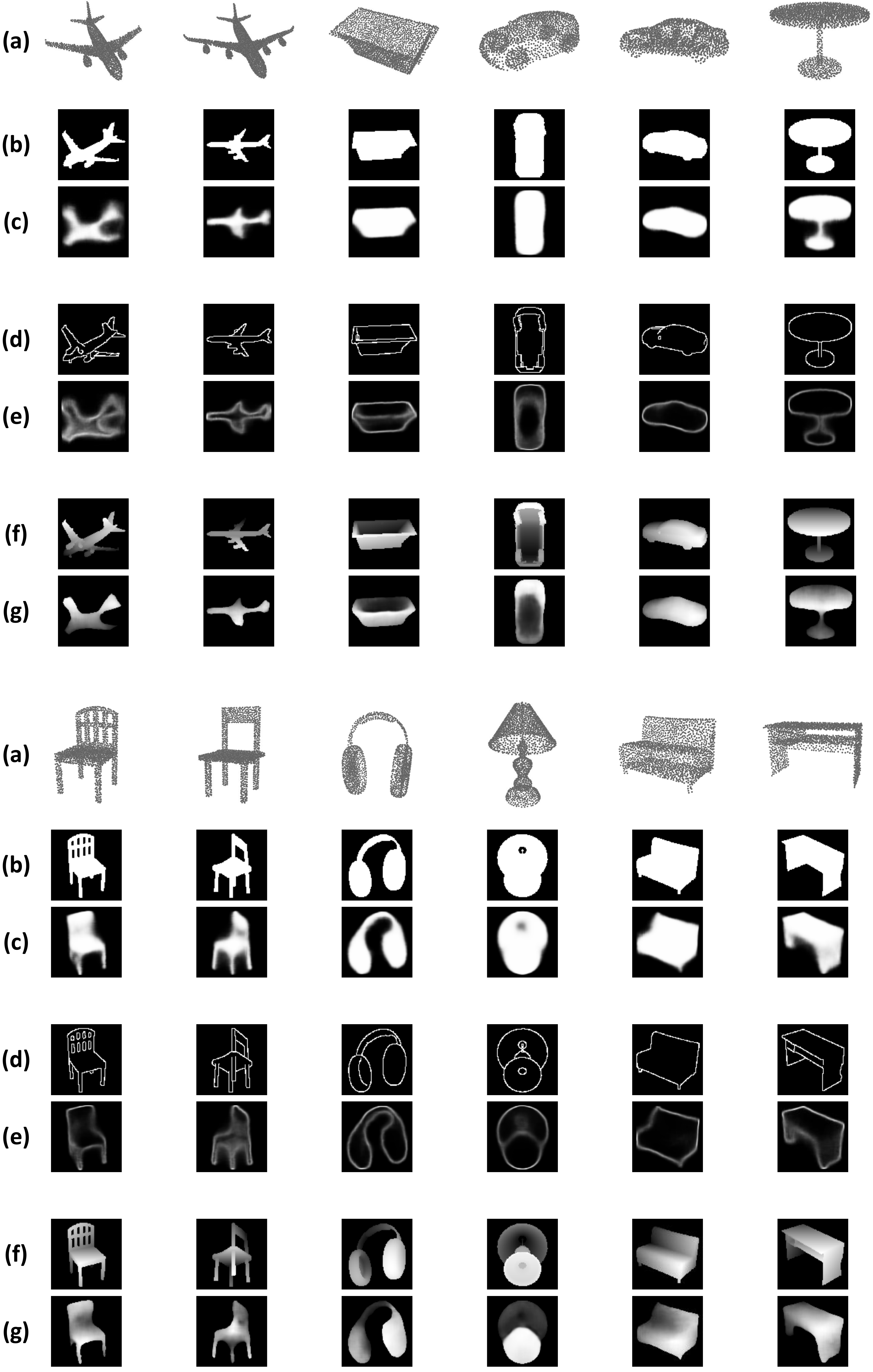}
	\caption{\textbf{Visual examples of our translated images.} Given (a) input point clouds, we compare ground-truth ((b), (d)) (f)) and translated ((c), (e), (g)) silhouette, contour, and depth images with respect to specified viewpoints.}
	\label{fig:trans-images-visual-examples}
\end{figure}

\begin{enumerate}
	\item \textbf{\textit{Prediction Results of Point-Wise Visibility Scores}.} In our proposed AVS-Pool, we tend to explicitly predict the positive scoring vector $\mathbf{S_v}$ that indicates point-wise visibility status with respect to the specified viewpoint. We performed quantitative evaluations by binarizing the predicted $\mathbf{S_v}$ with appropriate thresholds and compared it with the corresponding ground-truth mask $\mathbf{S^{\prime}_v}$. As reported in Figure~\ref{fig:visibility-accuracy-bar}, the overall accuracy of the predicted $\mathbf{S^{\prime}_v}$ reaches over $94\%$, meaning that the corresponding learnable layers are effectively optimized to provide reliable per-point visibility information. Note that, as demonstrated in Table~\ref{tab:ablation--avs-pool}, the task of visibility prediction itself is conducive to the learning effects of the overall pre-training pipeline. Besides, in Figure~\ref{fig:visibility-visual-examples}, we provided typical visual examples by simultaneously presenting the complete point clouds, the predicted and the ground-truth subsets of visible points (with respect to different randomly specified viewpoints).
	\item \textbf{\textit{Translation Results of 2D Rendered Images}.} In our implementation, we included three different forms of image translation objectives (single-channel images), i.e., depth $\mathbf{I}_d$, silhouette $\mathbf{I}_s$, and contour $\mathbf{I}_c$. In Figure~\ref{fig:trans-images-visual-examples}, we provided typical visual examples of our translated results and their corresponding ground-truth images (i.e., $\mathbf{I}^{\prime}_d$, $\mathbf{I}^{\prime}_s$, and $\mathbf{I}^{\prime}_c$) (with respect to different randomly specified viewpoints).
\end{enumerate}

Generally, it is observed that PointVST is able to produce reasonable view-specific image translation results. Still, since recovering fine-grained geometric details in the 2D image domain from sparse 3D point clouds is highly difficult, the resulting translation results show relatively coarse patterns. For example, it can hardly capture complex object contours, and the translated depth distributions are also less accurate in complex regions.

\subsection{Discussions} \label{sec:disc}

As the very first attempt towards a new design paradigm of self-supervised point cloud pre-training driven by cross-modal translation, our ultimate goal, as well as the core contribution, is to \textit{verify the viability} and \textit{explore the potential} of such direction. And through extensive experiments, the proposed PointVST did consistently show prominent performance superiority, despite the concise technical implementation without sophisticated learning mechanisms. 

Particularly, it is worth noting that our method is essentially different from neural rendering approaches \cite{xu2022point,metzer2022z2p,hu2023point2pix}, because the quality of the output images (e.g., being photo-realistic) is not a key consideration factor. The value of our PointVST, as a self-supervised learning framework for backbone pre-training, should be assessed by specific downstream task performances.

Finally, to better motivate the subsequent explorations along the proposed direction, below we discuss typical limitations of our current framework and also some promising future efforts. \\

\noindent \textbf{\textit{Limitations and extensions}.} The current technical implementation is designed to infer geometric appearance rendered from an outer viewpoint, which can cause information loss for point clouds that suffer from severe self-occlusion or have rich spatial structures inside the underlying surface. A more flexible or adaptively-learned viewpoint specification strategy is needed to relieve this issue. Besides, it might be indirect to acquire ground-truth 2D images corresponding to 3D point clouds with \textit{extremely sparse/noisy} distributions. Under such circumstances, the rendering results can be unsatisfactory, and thus we might have to switch to using realistic RGB camera images instead of synthetic ground-truth images. As a much more sophisticated problem setting suffering from multiple-solution phenomenon (i.e., one point cloud can correspond to varying plausible RGB appearance), this can be left for future studies. Another promising direction is to explore more advanced and effective ways of deducing view-specific feature representations, such as introducing attention mechanisms for fusing features of visible points.

\section{Conclusion} \label{sec:conclusion}

This paper aims at designing self-supervised pre-training frameworks for learning highly expressive and transferable geometric feature representations from unlabeled 3D point clouds. The core contribution is a new cross-modal translative pretext task, namely PointVST, posing a paradigm shift compared with previous generative and contrastive counterparts. Through extensive experiments on diverse evaluation protocols and downstream task scenarios, PointVST shows state-of-the-art performances with consistent and prominent gains. Our highly encouraging results indicate a much more promising way of integrating 2D visual signals into the pre-training pipelines of 3D geometric signals.



\quad

\quad

\quad

\quad

\begin{IEEEbiography}[{\includegraphics[width=1in,height=1.25in,clip,keepaspectratio]{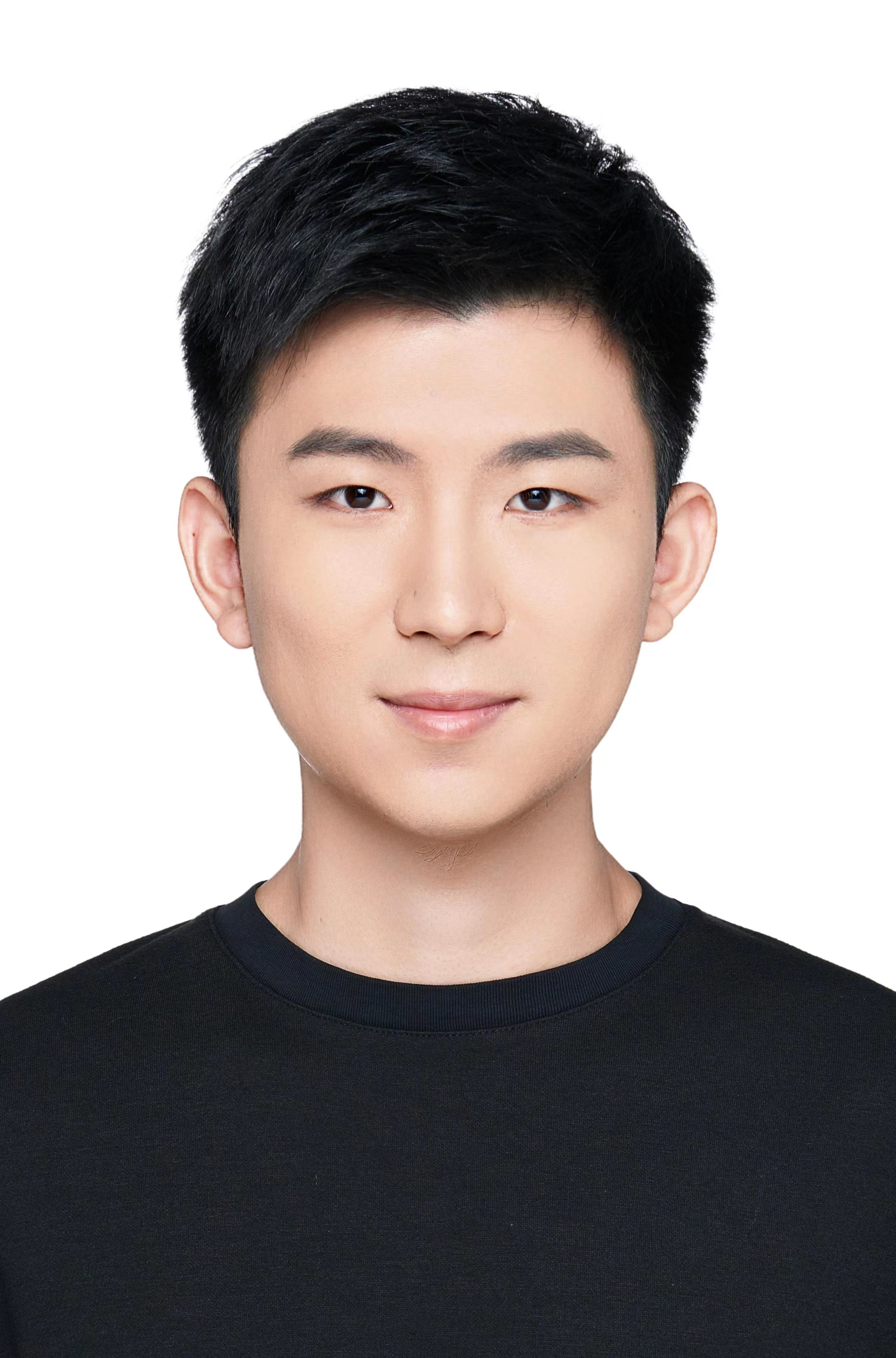}}]{Qijian Zhang}
	received the B.S. degree in Electronic Information Science and Technology from Beijing Normal University, Beijing, China, in 2019. Currently, he is a Ph.D. student (2020-present) under the Department of Computer Science, City University of Hong Kong, HKSAR. His research focuses on deep learning-based 3D point cloud processing, geometric computing, and cross-modal learning.
\end{IEEEbiography}

\begin{IEEEbiography}[{\includegraphics[width=1in,height=1.25in,clip,keepaspectratio]{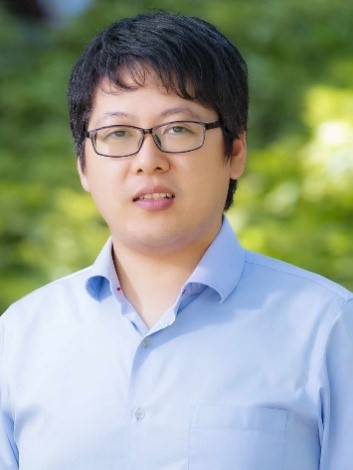}}]{Junhui Hou}
	is an Associate Professor with the Department of Computer Science, City University of Hong Kong.  He holds a B.Eng. degree in information engineering (Talented Students Program) from the South China University of Technology, Guangzhou, China (2009), an M.Eng. degree in signal and information processing from Northwestern Polytechnical University, Xi’an, China (2012), and a Ph.D. degree from the School of Electrical and Electronic Engineering, Nanyang Technological University, Singapore (2016). His research interests are multi-dimensional visual computing.
	
	Dr. Hou received the Early Career Award (3/381) from the Hong Kong Research Grants Council in 2018. He is an elected member of IEEE MSA-TC, VSPC-TC, and MMSP-TC. He is currently serving as an Associate Editor for \textit{IEEE Transactions on Visualization and Computer Graphics}, \textit{IEEE Transactions on Circuits and Systems for Video Technology}, \textit{IEEE Transactions on Image Processing}, \textit{Signal Processing: Image Communication}, and \textit{The Visual Computer}.
\end{IEEEbiography}

\vfill

\end{document}